\documentclass{egpubl}
\usepackage{pg2022}
 
\SpecialIssuePaper         % uncomment for final version of Computer Graphics Forum, special issue
\usepackage[T1]{fontenc}
\usepackage{dfadobe}  

\usepackage{cite}  % comment out for biblatex with backend=biber
\BibtexOrBiblatex
\electronicVersion
\PrintedOrElectronic
\ifpdf \usepackage[pdftex]{graphicx} \pdfcompresslevel=9
\else \usepackage[dvips]{graphicx} \fi

\usepackage{egweblnk} 
\usepackage{xspace}
\title{Implicit Neural Deformation for Sparse-View Face Reconstruction}

\author[Li et al.]
{\parbox{\textwidth}{\centering Moran Li$^{1,2}$\orcid{0000-0002-8149-1117},
      Haibin Huang\thanks{Corresponding author: jackiehuanghaibin@gmail.com}$^{1}$\orcid{0000-0002-7787-6428},
      Yi Zheng$^{1}$\orcid{0000-0002-6107-1836},
      Mengtian Li$^{1}$\orcid{0000-0001-6724-6177},
      Nong Sang$^{2}$\orcid{0000-0002-9167-1496},
      and Chongyang Ma$^{1}$\orcid{0000-0002-8243-9513}
        }
        \\
{\parbox{\textwidth}{\centering $^1$Kuaishou Technology, China \hspace{0.3in}
        $^2$Huazhong University of Science and Technology, China
      }
}
}
\usepackage[utf8]{inputenc} % allow utf-8 input
\usepackage[T1]{fontenc}    % use 8-bit T1 fonts
\usepackage{hyperref}       % hyperlinks
\usepackage{url}            % simple URL typesetting
\usepackage{booktabs}       % professional-quality tables
\usepackage{amsfonts}       % blackboard math symbols
\usepackage{nicefrac}       % compact symbols for 1/2, etc.
\usepackage{microtype}      % microtypography
\usepackage{graphicx}
\usepackage{subfig}
\usepackage{setspace}
\usepackage{multicol}
\usepackage{multirow}
\usepackage{makecell}
\usepackage{amsmath}
\usepackage{amssymb}
\usepackage{bbm}
\usepackage{hyperref}
\usepackage{color}
\usepackage{bbding}
\usepackage{microtype}      % microtypography
\usepackage{xcolor}         % colors
\usepackage{wrapfig}
\usepackage{enumitem}
\usepackage{pifont}
\usepackage{adjustbox}

\definecolor{RevisionColor}{rgb}{0,0.7,0}
\newcommand{\revision}[1]{#1}

\newcommand{\figtext}[1]{{\footnotesize #1}}

\renewcommand{\paragraph}[1]{\noindent \textbf{#1}}

\newcommand{\eg}{\emph{e.g.}}
\newcommand{\ie}{\emph{i.e.}}

\newcommand{\norm}[1]{\left\lVert#1\right\rVert}

\DeclareMathOperator*{\argmin}{arg\,min}

\newcommand{\loss}{\mathcal{L}}

\newcommand{\lossgeo}{\loss_\text{geo}}
\newcommand{\losseik}{\loss_\text{eik}}
\newcommand{\losspnt}{\loss_\text{I}}
\newcommand{\lossdeform}{\loss_\text{d}}
\newcommand{\losscode}{\loss_\text{reg}}
\newcommand{\losssw}{\loss_\text{sw}}
\newcommand{\lossrender}{\loss_\text{render}}
\newcommand{\lossrgb}{\loss_\text{rgb}}
\newcommand{\lossrgbswitch}{\loss_\text{rgb, sw}}
\newcommand{\lossmaskswitch}{\loss_\text{mask, sw}}
\newcommand{\lossmask}{\loss_\text{mask}}
\newcommand{\lw}{\lambda}
\newcommand{\lwpnt}{\lw_\text{I}}
\newcommand{\lweik}{\lw_\text{eik}}
\newcommand{\lwdeform}{\lw_\text{d}}
\newcommand{\lwreg}{\lw_\text{reg}}
\newcommand{\lwnormal}{\lw_\text{n}}
\newcommand{\lwr}{\tau}
\newcommand{\lwrrgb}{\lwr_\text{rgb}}
\newcommand{\lwrmask}{\lwr_\text{mask}}
\newcommand{\lwreik}{\lwr_\text{eik}}
\newcommand{\lwrdeform}{\lwr_\text{d}}
\newcommand{\lwrreg}{\lwr_\text{reg}}
\newcommand{\onsurfaceset}{\Omega_{I}}
\newcommand{\eikset}{\Omega_{D}}

\newcommand{\network}{f}
\newcommand{\geonetwork}{\network_\text{geo}}
\newcommand{\refnetwork}{\network_\text{ref}}
\newcommand{\deformnetwork}{\network_\text{deform}}
\newcommand{\rendernetwork}{\network_\text{render}}

\newcommand{\geonet}{Geometry Network\xspace}
\newcommand{\refnet}{Reference Network\xspace}
\newcommand{\deformnet}{Deformation Network\xspace}
\newcommand{\rendernet}{Rendering Network\xspace}

\newcommand{\parameters}{\theta}
\newcommand{\geoparameters}{\parameters_\text{geo}}

\newcommand{\surface}{S}
\newcommand{\distance}{s}
\newcommand{\position}{\mathbf{x}}
\newcommand{\offset}{\Delta\position}
\newcommand{\texture}{\mathbf{c}}
\newcommand{\normal}{\mathbf{n}}
\newcommand{\normalgt}{\hat{\normal}}

\newcommand{\geo}{\text{geo}}
\newcommand{\expsubsyb}{\text{exp}}
\newcommand{\idsubsyb}{\text{id}}
\newcommand{\latentcode}{\mathbf{z}}
\newcommand{\geometrycode}{\latentcode_\geo}
\newcommand{\geometrycodeswitch}{\latentcode_{\geo, \text{sw}}}
\newcommand{\expressioncode}{\latentcode_\expsubsyb}
\newcommand{\identitycode}{\latentcode_\idsubsyb}

\newcommand{\colorcode}{\latentcode_\texture}

\newcommand{\colorcodewresi}{\Tilde{\latentcode}_\texture}
\newcommand{\geometrycodewresi}{\Tilde{\latentcode}_\geo}

\newcommand{\colormean}{\bar{\latentcode}_\texture}
\newcommand{\geometrymean}{\bar{\latentcode}_\geo}
\newcommand{\geometrycoderesidual}{r_\geo}
\newcommand{\colorcoderesidual}{r_\texture}
\newcommand{\basis}{B}
\newcommand{\colorbase}{\basis_\texture}

\newcommand{\geometrybase}{\basis_\geo}
\newcommand{\codeweight}{W}

\newcommand{\cw}{W_\texture}
\newcommand{\geometryw}{\codeweight_\text{geo}}
\newcommand{\onsurface}{\position}
\newcommand{\geofeat}{l_\geo}

\newcommand{\viewdirection}{\mathbf{v}}
\newcommand{\cameracenter}{C_0}

\newcommand{\pixel}{\mathbf{p}}
\newcommand{\pixelset}{\mathcal{P}}
\newcommand{\rgbval}{\texture}
\newcommand{\rgbvalpred}{\rgbval}
\newcommand{\rgbvalgt}{\hat{\rgbval}}

\newcommand{\maskval}{m}
\newcommand{\maskvalpred}{\maskval}
\newcommand{\maskvalgt}{\hat{\maskval}}

\newcommand{\raystep}{t}

\newcommand{\cameraintri}{{\mathbf{K}}}
\newcommand{\cameraextri}{{\mathbf{R}}}

\newcommand{\realnumber}{\mathbb{R}}

\definecolor{amethyst}{rgb}{0.6, 0.4, 0.8}

\newcommand{\positionencoding}[1]{w^{#1}(\cdot)}

\begin{document}

% \inputfigure{fig_sgp_teaser}

\maketitle

\begin{abstract}
In this work, we present a new method for 3D face reconstruction from sparse-view RGB images. Unlike previous methods which are built upon 3D morphable models (3DMMs) with limited details, we leverage an implicit representation to encode rich geometric features. Our overall pipeline consists of two major components, including a geometry network, which learns a deformable neural signed distance function (SDF) as the 3D face representation, and a rendering network, which learns to render on-surface points of the neural SDF to match the input images via self-supervised optimization. To handle in-the-wild sparse-view input of the same target with different expressions at test time, we propose residual latent code to effectively expand the shape space of the learned implicit face representation as well as a novel view-switch loss to enforce consistency among different views. Our experimental results on several benchmark datasets demonstrate that our approach outperforms alternative baselines and achieves superior face reconstruction results compared to state-of-the-art methods.

\begin{CCSXML}
<ccs2012>
<concept>
<concept_id>10010147.10010371.10010396.10010397</concept_id>
<concept_desc>Computing methodologies~Mesh models</concept_desc>
<concept_significance>500</concept_significance>
</concept>
<concept>
<concept_id>10010147.10010371.10010396.10010402</concept_id>
<concept_desc>Computing methodologies~Shape analysis</concept_desc>
<concept_significance>500</concept_significance>
</concept>
</ccs2012>
\end{CCSXML}

\ccsdesc[500]{Computing methodologies~Mesh models}
\ccsdesc[500]{Computing methodologies~Shape analysis}

\printccsdesc
\end{abstract}

\section{Introduction}
\label{Section:Introduction}

In this paper, we tackle the problem of 3D face reconstruction given sparse-view input, \ie, to generate a textured face mesh based on a set of RGB images taken from different views. This problem is long-standing in both computer vision and computer graphics with many real-world applications, such as portrait manipulation and augmented/virtual reality.

Compared with reconstruction from a single RGB image or RGBD input, multi-view face reconstruction is a more practical setting with recent development of mobile devices, since it does not require additional depth senor but still provides rich information from different views about the target.
Previous methods \cite{beeler2010highqface_calib_mv, bradley2010high_calib_mv} propose to reconstruct 3D faces under controlled environments, where the multi-view images are captured from well-calibrated camera arrays with fixed lighting. Although these methods can successfully produce high-fidelity 3D face models, their usage scenarios are quite limited due to the complex hardware setup and their performances downgrade significantly for general input.
To address these drawbacks, some recent approaches \cite{bai20dfnrmvs,bai2021riggable,shang2020MCGNet_weak_super_mv} exploit 3DMMs \cite{blanz2003faceinitial_bfm,paysan20093dextendexpression_BFM,zhu2015high_extendpose_bfm} together with multi-view algorithms to leverage cross-view geometric consistency and demonstrate promising improvement.
However, those methods are built upon 3DMMs or the variants, where the number of vertices is limited and the topology is fixed.
Therefore, it remains challenging to generate a faithful 3D face with high-quality details from multi-view input, especially in an uncontrolled setting or when the input views are of different expressions.

\newcommand\teasergalleryfigurewidth{0.24}
\begin{figure}
\centering
\begin{minipage}{\linewidth}
\centering
\includegraphics[width=\linewidth]{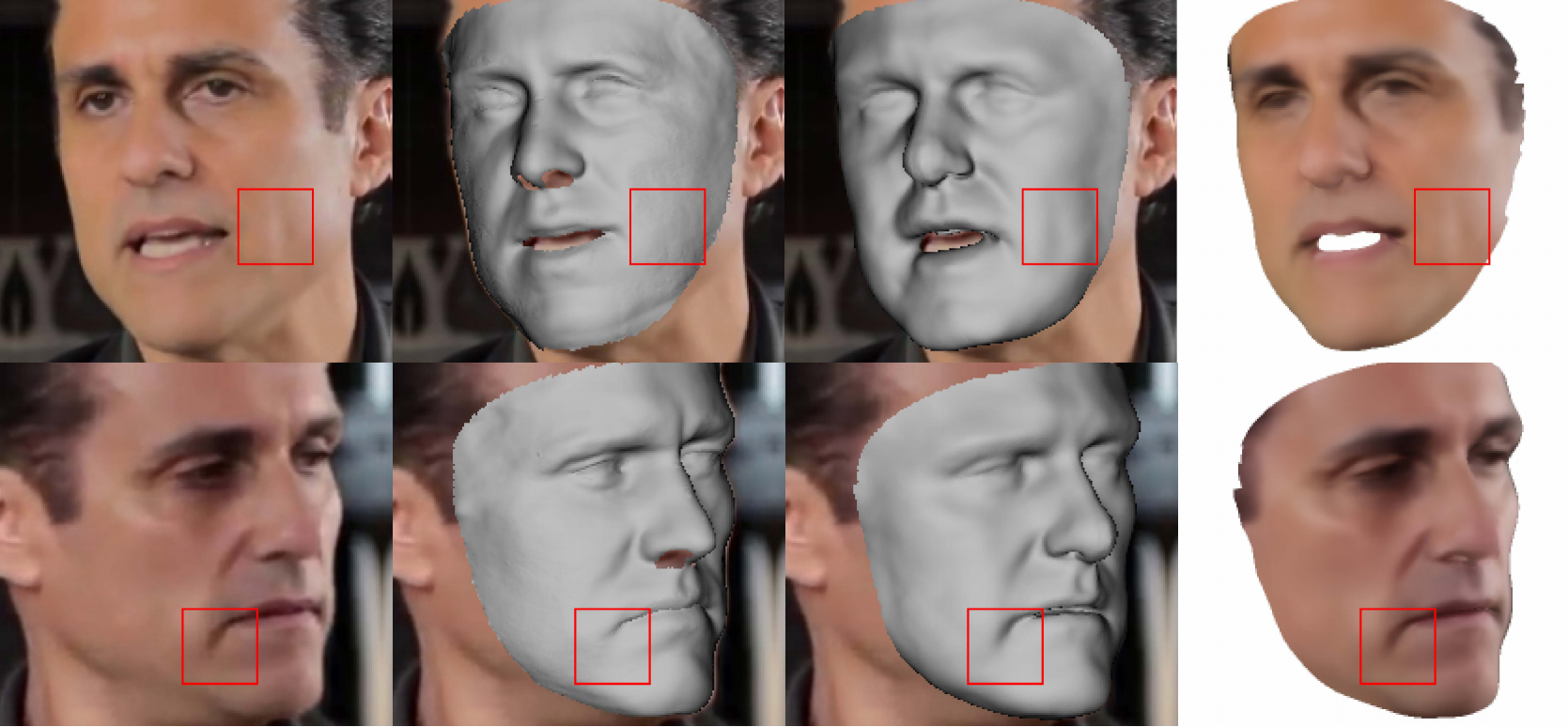}

\begin{minipage}{\teasergalleryfigurewidth\linewidth}
    \centering
    \figtext{Input}
\end{minipage}
\begin{minipage}{\teasergalleryfigurewidth\linewidth}
    \centering
    \figtext{DFNRMVS}
\end{minipage}
\begin{minipage}{\teasergalleryfigurewidth\linewidth}
    \centering
    \figtext{Ours}
\end{minipage}
\begin{minipage}{\teasergalleryfigurewidth\linewidth}
    \centering
    \figtext{Ours w/ texture}
\end{minipage}
\end{minipage}
\caption{Face reconstruction results of DFNRMVS \cite{bai20dfnrmvs} and our method from two-view inputs. The last two columns are our reconstructed geometry and the rendering results. Our method captures more local details (red boxes).
}
\label{fig:intro_teaser}
\end{figure}

In this work, our focus is to improve the generalization performance as well as the quality of sparse-view 3D face reconstruction by learning an implicit neural representation.
Our key insight is that, unlike 3DMMs that are limited by a pre-defined shape space, implicit functions such as SDFs can represent surfaces with arbitrary resolution and topology \cite{zheng2021deep,idr,igr,park2019deepsdf}.
To this end, we propose to learn a \geonet that serves as a neural SDF for reconstruction of the target 3D face.
Specifically, the proposed \geonet consists of two sub-modules, \ie, a \emph{\refnet} and a \emph{\deformnet}.
The \refnet is trained offline to learn the SDF of a mean face given the training set and provides an initiation of SDFs for optimization at the test time.
The \deformnet generates local details and changes topology if necessary by learning to deform the SDFs.
Our experiments show that such a decomposition effectively leverages a 3D face prior to enhance the generalization capability of the network and prevents the neural SDFs from collapsing or distorting during optimization with limited views (\eg, 2$\sim$4 views).

Inspired by \cite{idr}, we further present a \rendernet based on self-supervised optimization. This module learns to render on-surface points sampled from the implicit 3D face geometry. The self-supervision is achieved by minimizing the difference between rendered colors and the corresponding input images.
Additionally, we use geometry and color latent codes to encode shape and texture information among different instances to enhance the generalization ability of the trained network.
To expand the shape space of the learned neural SDF, we introduce \emph{residual latent code} at test time.
Furthermore, we design a \emph{view-switch loss} via exchanging the latent code among different views and minimizing the rendering loss to enforce consistency across different views.
As a result, our method can reconstruct 3D faces from sparse-view input with high-fidelity details. See Figure~\ref{fig:intro_teaser} for an example of 3D face reconstruction from two in-the-wild images of the same person but with different expressions.

To summarize, our main contributions are as follows:
\begin{itemize}%[leftmargin=*]
\item We present a pipeline for 3D face reconstruction from sparse-view input, including a \geonet to learn a deformable implicit neural representation for 3D shapes and a \rendernet to model the facial texture.

\item We propose a novel view-switch loss as well as a newly designed latent code space of the implicit morphable model. These two terms help expand the underlying shape space and enforce cross-view consistency at the test time.

\item We conduct both qualitative and quantitative evaluations on benchmark datasets to demonstrate that our method outperforms baseline approaches and is comparable to state-of-the-art face reconstruction algorithms.

\end{itemize}

\section{Related Work}
\label{Section:Related:Work}

The literature on 3D face reconstruction is vast and the algorithm input ranges from depth map \cite{kazemi2014realface_depthmap} and single image \cite{tuan2017faceregressing_sv,kim2018inversefacenet_sv,richardson2017learning_bfmdetails_sv,dou2017end_learnbfmfeasible_sv,zhu2020reda_facerender_sv,chen2019photo_sv,feng2018joint_sv,shang2020MCGNet_weak_super_mv,chenDeepFaceEditing2021,xu2020deep,gao2020portrait} to multi-view images \cite{bai2021riggable,bai20dfnrmvs, wu2019mvf_weak_super_mv,deng2019accurate_weak-super_mv,sanyal2019learning_weak_super_mv,chan2021efficient} and videos \cite{garrido2016reconstruction_video, tewari2019fml_weak_super_video_mv}.
Since our main focus is 3D face reconstruction from sparse-view images using neural SDFs as the geometric representation, in this section, we briefly review 3D morphable models, multi-view 3D face reconstruction methods, and the most relevant implicit neural representations.

\paragraph{Face morphable models.}
The well-known 3D morphable model \cite{blanz2003faceinitial_bfm,paysan20093dextendexpression_BFM,cao2013facewarehouse_bfm,zhu2015high_extendpose_bfm} is a bilinear parametric method that decomposes the face geometry{\slash}texture into a template and a deformation component with respect to this template based on principal component analysis (PCA).
Due to their simplicity and effectiveness, 3DMMs are widely used in faces reconstruction and animation.
However, the capability of such models is limited by the basis of PCA. Even though several recent methods (\eg, \cite{cao2013facewarehouse_bfm, yang2020facescape_bfm, smith2020morphable_extend_bfm}) propose to extend the face basis with more 3D face scans from larger datasets, the geometry or texture space of those methods is still a subspace of real-world face space. For a complete report of 3D morphable face, we refer to \cite{20203dMMsummary}.

\paragraph{Multi-view face reconstruction.}
Existing learning-based algorithms for multi-view 3D face reconstruction can be roughly categorized into supervised methods \cite{garrido2016reconstruction_video,cao2014displaced_video,bai20dfnrmvs} and self-supervised methods \cite{deng2019accurate_weak-super_mv,sanyal2019learning_weak_super_mv,wu2019mvf_weak_super_mv}.
\cite{garrido2016reconstruction_video} exploits parametric geometry prior information to learn a plausible coarse face mesh and fine-scale details are captured via shading-based refinement from videos. \cite{bai20dfnrmvs} proposes to expand the basis of 3DMMs via adaptive optimization to improve the representation of such parametric models and enforce multi-view consistency.
To alleviate the requirement of large-scale 3D scan datasets, some researchers tackle this problem in a self-supervised manner. \cite{deng2019accurate_weak-super_mv} uses aggregated complementary information among different images to achieve multi-view reconstruction. However, those models are built upon 3DMMs, where the mesh topology is fixed and cannot represent high-frequency details easily.

\paragraph{Implicit neural representation.}
In recent years, methods based on implicit neural representations are emerging for shapes \cite{deng2021deformed,yariv2021volume,park2019deepsdf, atzmon2020sal_object_sdf, davies2020overfit_object_sdf, genova2020local_sdf,takikawa2021nvidia_lod_sdf,igr, zheng2021deep,tang2021octfield} and scenes \cite{eslami2018_scene_sdf,sitzmann2019scene_sdf,jiang2020local_scene_sdf,kohli2020semantic_scene_sdf}. 
The seminal work DeepSDF \cite{park2019deepsdf} encodes a category of shapes into a neural network, while the specific features of each instance are encoded into a latent code. 
Based on DeepSDF, \cite{duan2020curriculum_sdf} proposes a curriculum architecture to enhance the quality of the reconstructed shape. Those methods are used to obtain the implicit neural representations of shapes, objects, or scenes with 3D data (\eg, point cloud) as the supervision.
\cite{zheng2021deep,deng2021deformed} further decompose the implicit neural representation for 3D geometry into a deformation and a template implicit representation.  
\cite{saito2019pifu_sdf} exploits pixel-aligned implicit function to estimate the surface of human subjects and the corresponding texture.

Recently, \cite{sun2021nelf, gafni2021dynamic, mildenhall2020nerf_sdf, yu2020pixelnerf_sdf, kellnhofer2021neuralumigraph_render_sdf} propose to synthesize novel views from a set of images by reconstructing the underlying 3D scene{\slash}object geometry and the neural radiance field at the same time.
\cite{idr, wang2021neus} use neural SDF to represent surface geometry and reconstruct the target shape from multi-view images.
However, most of those techniques require more than 30 images from different viewpoints for each object/scene, and the reconstructed surface will collapse for sparse-view input due to the lack of prior information about the object/scene. 
\revision{Among those methods, the most relevant one is i3DMM~\cite{yenamandra2020i3dmm}, which builds implicit 3D morphable models for human heads with hair from a dataset of 3D scans and requires calibrated dense-view capture of the subject.
Our goal instead is to reconstruct a 3D face from sparse-view 2D input and thus is more challenging than the setting of i3DMM.
}%revision

\begin{figure*}%[ht]
\begin{center}
\includegraphics[width=0.97\linewidth]{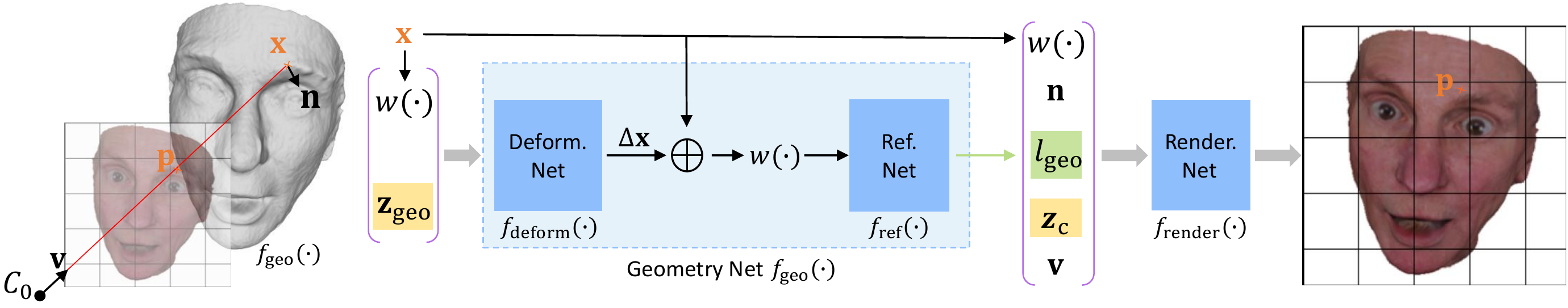}
\end{center}
\caption{Overview of the proposed method. Our method consists of the \emph{Geometry Network} and \emph{Rendering Network}. And the \emph{Geometry Network} can be decomposed into a \emph{Deformation Network} and a \emph{Reference Network}. $\viewdirection$ is the view direction from the camera center $\cameracenter$ to the randomly sampled pixel $\pixel$. $\onsurface$ is the corresponding on-surface point, and $\normal$ is the normal vector. $\geometrycode, \colorcode$ are the geometry and color latent codes, respectively. $\geofeat$ is the feature from Geometry Network (\ie, $\geonetwork(\cdot)$). $w(\cdot)$ is the positional encoding function. $\color{amethyst}\left(\; \right)$ denotes the concatenate operation. For conciseness, we only present one view of the instance, while most of our experiments are sparse-view inputs.}
\label{fig:overall}
\end{figure*}

\section{Our Method}

\subsection{Face Representation and Problem Statement}

In an implicit neural representation based on signed distance field (SDF), the geometry of a 3D face can be represented as the zero level set of a scalar valued network $\network$:
\begin{equation}
    \surface_{\parameters} = \big\{\position \in \realnumber^3 | \network_{\parameters}(\position) = 0 \big\}
\label{eqn:sdf}
\end{equation}
where the network $\network_{\parameters}(\position)$ gives the signed shortest distance $\distance$ of a query point $\position \in \realnumber^3$ to the face geometry $S_{\parameters}$ and $\parameters \in \realnumber^m$ are learnable parameters of the network.
To model the facial texture, we extend the network output to include a vector $\texture \in \realnumber^3$, which represents the color of the closest point on the face from the query point.

To further model various faces of different identities and expressions, we introduce a latent code $\latentcode$ to represent the face instance in a portrait image.
Following i3DMM~\cite{yenamandra2020i3dmm}, we denote the network as $\network_{\parameters}(\position, \, \latentcode)$ to take this latent code as additional input.

In both the training and test stages, we jointly optimize the network parameters $\parameters$ and the latent code $\latentcode$ as described in detail below, in order to obtain the desired morphable model and the corresponding implicit representation of each face instance.
To simplify the notation, we omit $\parameters$ in the subscript and rewrite the network as
\revision{$\{\distance, \, \texture \} = \network(\position, \, \latentcode)$}.

\subsection{Network Components}
As illustrated in Figure~\ref{fig:overall}, our overall framework consists of two network components, \ie, a \geonet $\geonetwork$ and a \rendernet $\rendernetwork$.
Accordingly, the latent code $\latentcode$ of each face instance can be decomposed into two parts, \ie, a geometry code $\geometrycode$ and a color code $\colorcode$, which are used as input of $\geonetwork$ and $\rendernetwork$, respectively.

\paragraph{\geonet.}
Our \geonet $\geonetwork$ is a scalar valued function to model the implicit 3D face shape.
We follow i3DMM~\cite{yenamandra2020i3dmm} to further decompose $\geonetwork$ into two successive components, \ie, a \refnet $\refnetwork$ to learn an implicit reference shape, and a \deformnet $\deformnetwork$ to predict a deformation offset $\offset$ conditioned on the reference shape.
The \refnet can be considered as a neural version of the mean face in traditional 3DMMs~\cite{blanz1999morphable_bfm}, while the \deformnet models the per-instance variations from the mean face.
As a result, the \geonet can be formulated as:
\revision{
\begin{equation}
\begin{aligned}
\geonetwork(\position, \geometrycode) &= \refnetwork \circ \deformnetwork(\position, \geometrycode) \\
& = \refnetwork(\position + \offset).
\label{eqn:geo_net_decomposition}
\end{aligned}
\end{equation}
}

\paragraph{\rendernet.} % Texture Network?
Our \rendernet $\rendernetwork$ is introduced to model the face texture in a self-supervised manner, where the texture information is encoded as the color latent code $\colorcode$ for each face instance.
Therefore, for a given surface point $\position$ of a certain instance, the RGB value $\texture(\position, \, \colorcode)$ can be modeled using our \rendernet by taking several factors into account together:
\begin{equation}
    \texture(\position, \, \colorcode) = \rendernetwork(\position, \colorcode, \normal, \viewdirection, \geofeat),
\label{eqn:rendering_network}
\end{equation}
where $\normal$ is the surface normal, $\viewdirection$ is the view direction, and $\geofeat \in \realnumber^{256}$ are geometric features computed as additional output by the \geonet $\geonetwork$.
Note that the \revision{normalized} gradient of the signed distance computed by the \geonet at a point $\position$ is the corresponding surface normal, \ie, \revision{$\normal=\nabla{\geonetwork(\position, \geometrycode)} / \norm{ \nabla{\geonetwork(\position, \geometrycode)} }_2$}.

\subsection{Network Training}

\paragraph{Dataset.}
We use a training partition of the Stirling{\slash}ESRC \cite{stir3d} dataset to train our network, which contains more than $700$ registered 3D face scans of about $95$ subjects.
To prepare training data for the \geonet $\geonetwork$, the 3D scans are scaled to fit into a unit bounding box and then aligned to the same orientation. We then randomly sample 860K on-surface points from each registered scan in a uniform distribution. 
We consider these points together with the corresponding normals as the zero level set of each SDF and use them to train the \geonet. To prepare training data for the \rendernet $\rendernetwork$, we render about $40$ RGB images of each 3D face scan from random view directions.
For each rendered image, we also compute a binary mask to represent the face region.

\paragraph{Geometry loss function.}
Given a face instance $i$ with a geometry latent code $\geometrycode^i$ and a set of sample points $\onsurfaceset$, the overall geometry loss function $\lossgeo$ is computed as:
\begin{equation}
\lossgeo(\geometrycode^i) = \lwpnt \losspnt + \lwdeform \lossdeform + \lweik \losseik + \lwreg \losscode
\label{eqn:loss_geo}
\end{equation}
where $\lwpnt$, $\lwdeform$, $\lweik$, and $\lwreg$ are hyperparameters to balance different loss terms. We set $\lwpnt=1$, $\lwdeform=0.01$, $\lweik=0.1$, and $\lwreg=\text{1e-4}$ in our experiments.

In Eq. \eqref{eqn:loss_geo}, $\losspnt$ is a reconstruction loss to enforce the signed distance values of sampled on-surface points are close to zero and the normals of those points are close to the ground truth values:
\revision{
\begin{equation}
\begin{aligned}
\losspnt = \frac{1}{\norm{ \onsurfaceset }_1} \sum_{ \onsurface_j \in \onsurfaceset} \big(& \norm{ \geonetwork(\onsurface_j, \geometrycode^i) }_1 \\
 & + \lwnormal \norm{\normal_j - \normalgt_j}_2\big),
\label{eqn:loss_pnts}
\end{aligned}
\end{equation}
where $\norm{\cdot}_1$ is the L1 norm, $\norm{\cdot}_2$ is the L2 norm,
}%revision
$\onsurfaceset = \{ \onsurface_j\}_{j \in I }$ is a randomly sampled set of the on-surface points, \revision{$\normal_j=\nabla{\geonetwork(\onsurface_j, \geometrycode^i)} / \norm{ \nabla{\geonetwork(\onsurface_j, \geometrycode^i)} }_2$}, and $\normalgt_j$ is the ground truth surface normal of the on-surface point $\onsurface_j$. We set $\lwnormal=1$ in our experiments.
$\lossdeform$ in Eq. \eqref{eqn:loss_geo} is the regularization of the deformation offset:
\begin{equation}
    \lossdeform = \frac{1}{\norm{\onsurfaceset}_1} \sum_{ \onsurface_j \in \onsurfaceset} \norm{\offset_j^i}_2  = \frac{1}{\norm{\onsurfaceset}_1} \sum_{ \onsurface_j \in \onsurfaceset} \norm{\deformnetwork(\onsurface_j, \geometrycode^i)}_2,
\label{eqn:loss_deform}
\end{equation}
and $\losscode=\norm{\geometrycode}_2$
is the regularization for the geometry latent code. Finally, $\losseik$ is the Eikonal term to avoid universe zero and ensures that $\geonetwork$ approximates valid SDFs \cite{igr,idr}:
\begin{equation}
\losseik = \frac{1}{\norm{\eikset}_1} \sum_{ \onsurface'_j \in \eikset}\big( \norm{\nabla{\geonetwork(\onsurface'_j, \geometrycode^i)}}_2 - 1\big)^2,
\label{eqn:loss_eik}
\end{equation}
where $\eikset = \{ \onsurface'_j \}_{j \in D}$ is a set of points sampled from a uniform distribution within a unit bounding box.

Given $N$ face instances within a mini-batch, we can jointly optimize the parameters $\geoparameters$ of the \geonet $\geonetwork$ and the geometry latent codes $\{ \geometrycode^i, i=1,\ldots,N \}$ of these $N$ instances by solving the optimization problem below:
\begin{equation}
\argmin_{\geoparameters, \, \{ \geometrycode^i \} } \frac{1}{N} \sum^N_{i=1} \lossgeo(\geometrycode^i)
\label{eqn: loss_igr_min}
\end{equation}

\paragraph{Rendering loss function.}
Given a pair of a rendered RGB image and the corresponding face mask, we randomly sample a subset of pixels $\pixelset$ in the image plane and use the following rendering loss function $\lossrender$ to train our \rendernet $\rendernetwork$:
\begin{equation}
\lossrender = \lwrrgb \lossrgb + \lwrmask \lossmask + \lwrdeform \lossdeform + \lwreik \losseik + \lwrreg \losscode'
\label{eqn:loss_render}
\end{equation}
where $\lwrrgb$, $\lwrmask$, $\lwrdeform$, $\lwreik$, and $\lwrreg$ are set to 1, 100, 1e-4, 0.01, and 1e-4 to balance different loss terms. $\losscode'=\norm{\geometrycode}_2 + \norm{\colorcode}_2$. In Eq.~\eqref{eqn:loss_render}, the loss terms $\lossdeform$ and $\losseik$ are similar to those in Eq.~\eqref{eqn:loss_geo}, while $\lossrgb$ is the RGB reconstruction loss and $\lossmask$ is the mask loss, respectively.

Specifically, the RGB reconstruction loss is computed as:
\begin{equation}
\lossrgb = \frac{1}{\norm{\pixelset}_1} \sum_{\pixel \in \pixelset} \norm{\rgbvalpred_\pixel - \rgbvalgt_\pixel}_1
\label{eqn:loss_rgb}
\end{equation}
where $\rgbvalpred_\pixel$ is the RGB value at the pixel $\pixel$ predicted by the \rendernet, and $\rgbvalgt_\pixel$ is the corresponding ground truth RGB value.
We use cross-entropy loss to compute the mask loss $\lossmask$ as below:
\begin{equation}
\lossmask = \frac{1}{\norm{\pixelset}_1} \sum_{\pixel \in \pixelset} \text{CE}(\maskvalpred_\pixel, \maskvalgt_\pixel
)
\label{eqn:maskloss}
\end{equation}
where $\maskvalpred_\pixel$ and $\maskvalgt_\pixel$ are the predicted and the ground truth mask values at the pixel $\pixel$, respectively.
As in \cite{idr}, we use a sigmoid function to achieve differentiable rendering of the mask.

\paragraph{Training strategy.}
Reconstructing 3D face geometry from a sparse set of RGB input (\ie, $2 \sim 4$ views) with various expressions is an ill-posed problem.
Besides, the proposed self-supervision is achieved by enforcing a similarity between the rendered RGB values and the ground truth RGB values.
As a result, the \rendernet tends to overfit the input RGB images and the implicit neural geometry may collapse.
To alleviate this problem, we first optimize the \geonet with the geometry loss $\lossgeo$ to obtain a good initialization. Then, we jointly optimize the \rendernet and the \geonet via the rendering loss $\lossrender$.

% % -----------------------------------
\begin{figure*}
\centering
\includegraphics[width=0.99\linewidth]{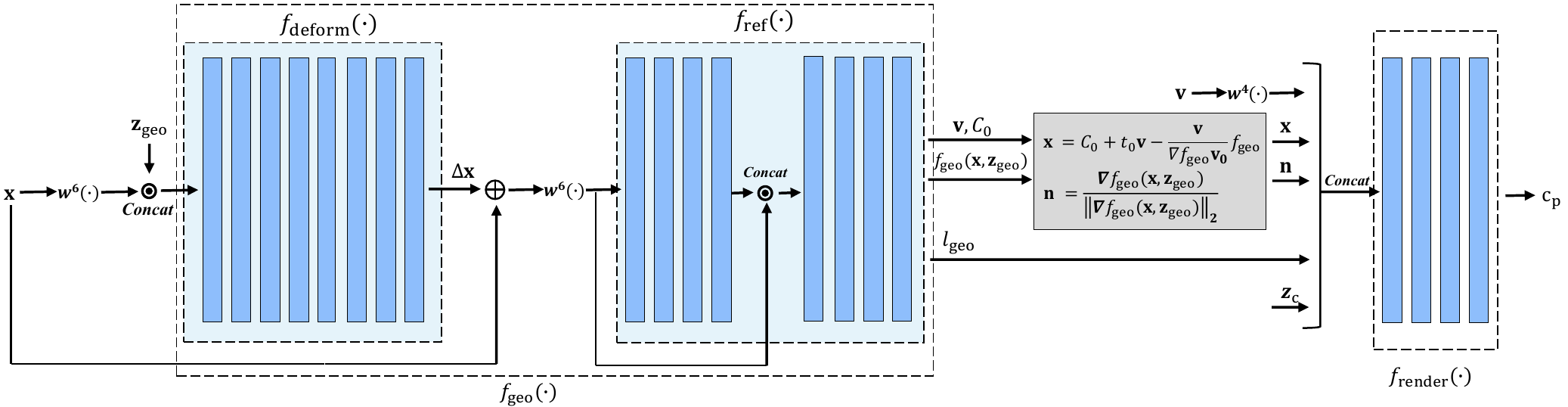}
\caption{Implementation details of our network architecture.
The geometry network $\geonetwork$ consists of a deformation network $\deformnetwork$ and a reference network $\refnetwork$.
$\geometrycode \in \mathbb{R}^{256}$ and $\colorcode \in \mathbb{R}^{128}$ are the geometry latent code and the color latent code, respectively. 
$\onsurface$ is the on-surface point of the corresponding image plane pixel $\pixel$.
$\rgbvalpred_\pixel \in \mathbb{R}^3$ is the predicted pixel RGB value.
We use $\positionencoding{6}$ and $\positionencoding{4}$ as the positional encoding functions for the on-surface point and the view direction, respectively. 
$\cameracenter \in \mathbb{R}^3$ is the camera center, and $\viewdirection \in \mathbb{R}^3$ is the view direction that crosses the pixel $\pixel$.  $\odot$ and $\oplus$ denote the \emph{concatenation} and \emph{add} operations, respectively.
Each blue box represents a fully connected layer.
}
\label{fig:supp_netarchi}
\end{figure*}

\subsection{Test-Time Reconstruction}
\paragraph{Estimation of camera parameters.}
To handle in-the-wild images at test time, we estimate camera parameters by optimizing the L1 distance between the projected on-surface point $\onsurface$ and the ground-truth pixel location $\pixel$ :
\begin{equation}
    \argmin_{\cameraintri, \cameraextri} \norm{\cameraintri \cameraextri \onsurface^{\top}_1 - \pixel}_1
\label{eqn:camera_param}
\end{equation}
where $\cameraintri \in \realnumber^{3\times3}$ are camera intrinsic parameters, $\cameraextri \in \realnumber^{3\times4}$ is the rotation matrix, $\onsurface_1 = [\onsurface, 1] \in \realnumber^{1\times4}$ are homogeneous coordinates of the point $\onsurface$\revision{, and the superscript notation $\top$ is the transpose operator}.
In our method, the on-surface point is defined as the first intersection point of the ray across the pixel $\pixel$ and the face geometry $S_{\parameters}$.
The intersection point can be represented as a differentiable function of the implicit geometry and camera parameters.
We use the differentiable sphere-tracing method \cite{liu2020dist} to find the on-surface point.

\revision{We use a coarse-to-fine strategy to estimate camera parameters.
Specifically, we first use the 68 landmarks of the mean face to obtain a coarse estimation of the camera parameters.
Then we refine the camera parameters based on Eq.~\eqref{eqn:camera_param} using the Adam optimizer with a learning rate of 1e-3.
}%revision

\paragraph{Residual latent code.}
Given a sparse set of RGB images with various expressions of an instance, it is difficult to learn the latent code directly using the rendering loss described above.
Hence, we use principal component analysis (PCA) on the learned latent codes of the training set and infer the weights of those PCA basis at test time.
Specifically, the latent code of an instance $i$ can be represented as the weighted sum of the principal components, such as:
\begin{equation}
\begin{aligned}
\geometrycode^{i} &= \geometrymean +\geometryw^{i} \geometrybase, \\
\colorcode^{i} &= \colormean + \cw^{i} \colorbase
\end{aligned}
\label{eqn:pca_latentcode}
\end{equation}
where $\geometryw^i \in \realnumber^{1 \times m_\geo}, \cw^i \in \realnumber^{1 \times m_\texture}$ are the weights to combine the basis of geometry and color latent codes, respectively.
$\geometrymean \in \realnumber^{d_\geo}$ and $\colormean \in \realnumber^{d_\texture}$ are the mean latent code for the identity and color.
$\geometrybase \in \realnumber^{m_\geo \times d_\geo}, \colorbase \in \realnumber^{m_\texture \times d_\texture}$ are the PCA basis of geometry and color latent code space.
$m_\geo$ and $m_\texture$ are the number of the principal components.
\revision{We set $m_\geo$ and $m_\texture$ to be 85 and 65, respectively, such that the explained variance is more than 96\% for latent space decomposition.}
At test time, we can obtain the combination weights via solving the optimization problem:
\begin{equation}
\argmin_{\geometryw^{i}, \cw^{i}} \frac{1}{N} \sum^N_{i=1} \lossrender(\geometrycode^{i}, \colorcode^{i})
\label{eqn:test_pca_latentcode_weight}
\end{equation}

We further introduce residual latent codes $\geometrycoderesidual^{i} \in \realnumber^{d_\geo}$ and $\colorcoderesidual^{i} \in \realnumber^{d_\texture}$ for each instance to expand the underlying representation space. Hence, the latent codes of an instance $i$ can be formulated as:
\begin{equation}
\begin{aligned}
\geometrycodewresi^{i} &= \geometrycode^{i} + \geometrycoderesidual^{i},\\
\colorcodewresi^{i} &= \colorcode^{i} + \colorcoderesidual^{i}
\end{aligned}
\label{eqn:pca_latentcode_extend}
\end{equation}
The optimization problem at test time can be formulated as:
\begin{equation}
\argmin_{\geometryw^{i}, \cw^{i}, \geometrycoderesidual^{i}, \colorcoderesidual^{i}} \frac{1}{N} \sum^N_{i=1} \lossrender(\geometrycodewresi^{i}, \colorcodewresi^{i}).
\label{eqn:test_pca_latentcode_weight_extend}
\end{equation}

\paragraph{View-switch loss.}
A key problem in multi-view reconstruction is how to enforce view consistency to better leverage the multi-view information.
In our method, the view consistency is enforced from two aspects.
First, we divide the geometry latent code into an identity component and an expression component (\ie, $\geometrycode = \{ \identitycode, \expressioncode \}$). Therefore, we can impose cross-view consistency implicitly via enforcing different views of the same instance to share the same identity and color latent code.
Second, we propose a view-switch loss $\losssw$ to further impose an explicit view consistency.
The key observation to inspire our view-switch loss is that the rendered images from different expression latent codes of the same identity under the same camera pose should be similar when ignoring local regions that are more likely to be influenced by varying expressions (\eg, the mouth region).

Specifically, given two views $i_m$ and $i_n$ from a sparse set of images for an instance $i$, we replace the expression latent code of the view $i_m$ with that of the view $i_n$ (\ie, $\expressioncode^{i_n}$). Then, we use the switched geometry latent code (\ie, $\geometrycodeswitch^{i_m} = \{ \identitycode^i, \expressioncode^{i_n} \}$) to calculate the RGB loss as Eq.~\eqref{eqn:loss_rgb} and the mask loss as Eq.~\eqref{eqn:maskloss}.
To reduce the impact of expression variations among different views, we use a face parsing network \cite{yu2018bisenet} to omit pixels in the mouth region.
We denote the RGB and mask losses after switch views and latent code as the view-switch loss $\losssw$ to distinguish from previous versions in Eq.~\eqref{eqn:loss_render}.
Hence, the total loss function $\lossrender'$ at test time is:
\begin{equation}
\begin{aligned}
\lossrender' &= \lossrender + \mu \losssw \\
          &= \lossrender + \mu (\lwrrgb \lossrgbswitch + \lwrmask \lossmaskswitch)
\label{eqn:loss_infer_wsw}
\end{aligned}
\end{equation}
where $\mu$, $\lwrrgb$, and $\lwrmask$ are set to 0.1, 1, and 100 in our experiments.

\paragraph{Mesh recovery.}
To recover the mesh from our neural SDF, \ie, $S^i_{\parameters}$ for a face instance $i$, we use the Marching Cubes algorithm \cite{lorensen1987marching} with a resolution of $250$, which is a trade-off setting to balance the output quality and the computational cost.

\section{Experiments}
\label{sec:Experiments}

\subsection{Implementation Details}
\label{subsec:details}

\newcommand\galleryfigurewidth{0.32}
\begin{figure*}%[t]
\centering
\subfloat[Instance 1]{\begin{minipage}{0.495\linewidth}
\includegraphics[width=\galleryfigurewidth\linewidth]{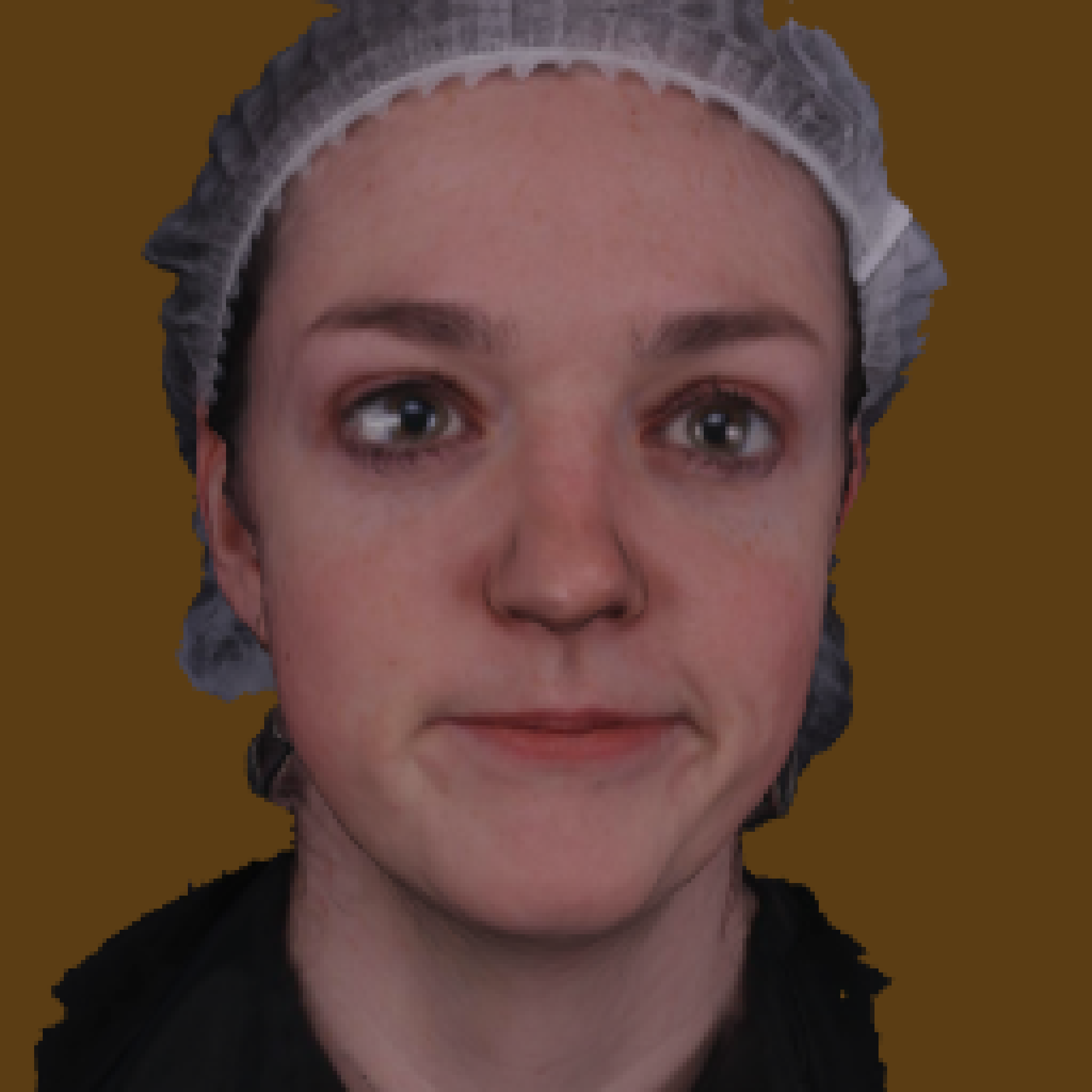}
\includegraphics[width=\galleryfigurewidth\linewidth]{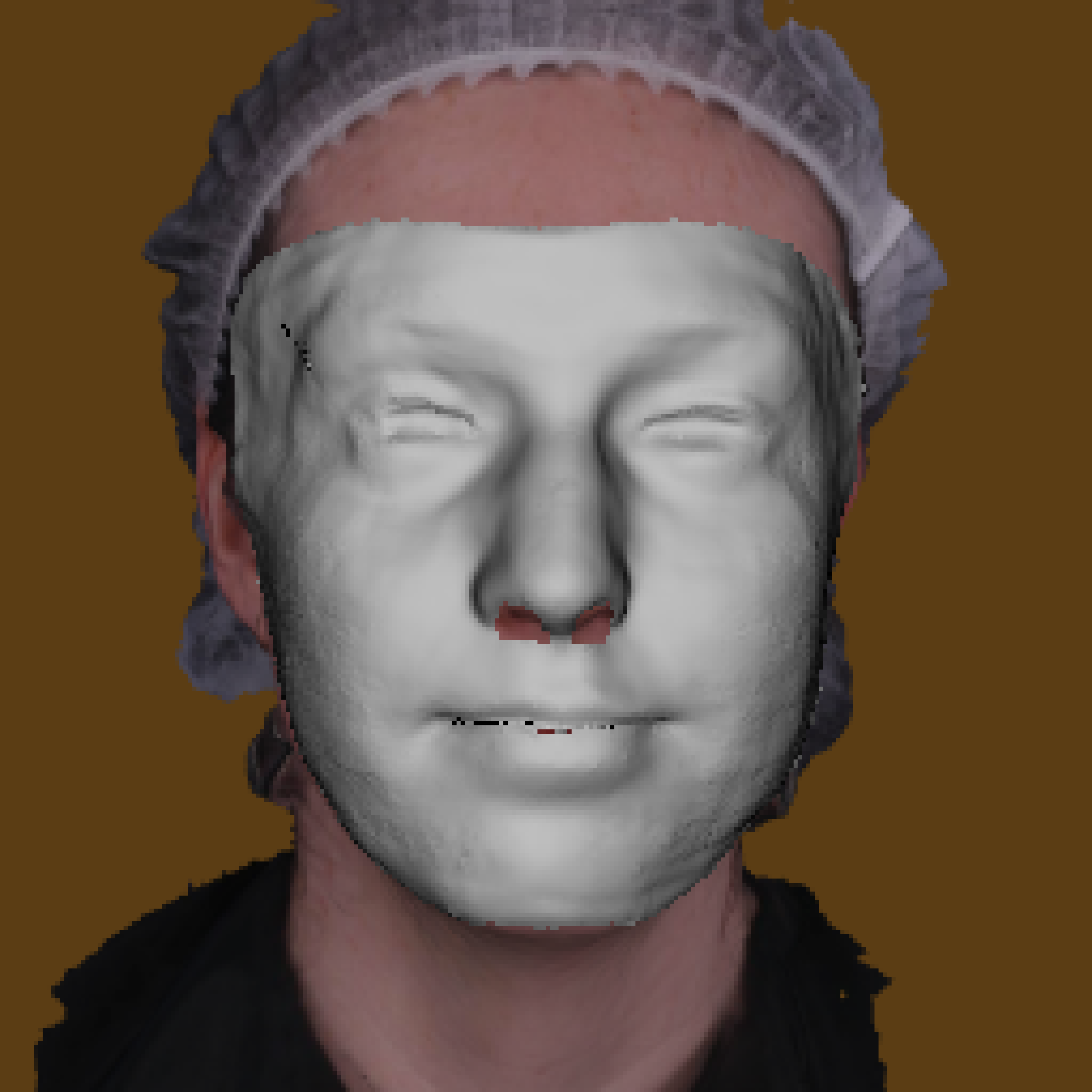}
\includegraphics[width=\galleryfigurewidth\linewidth]{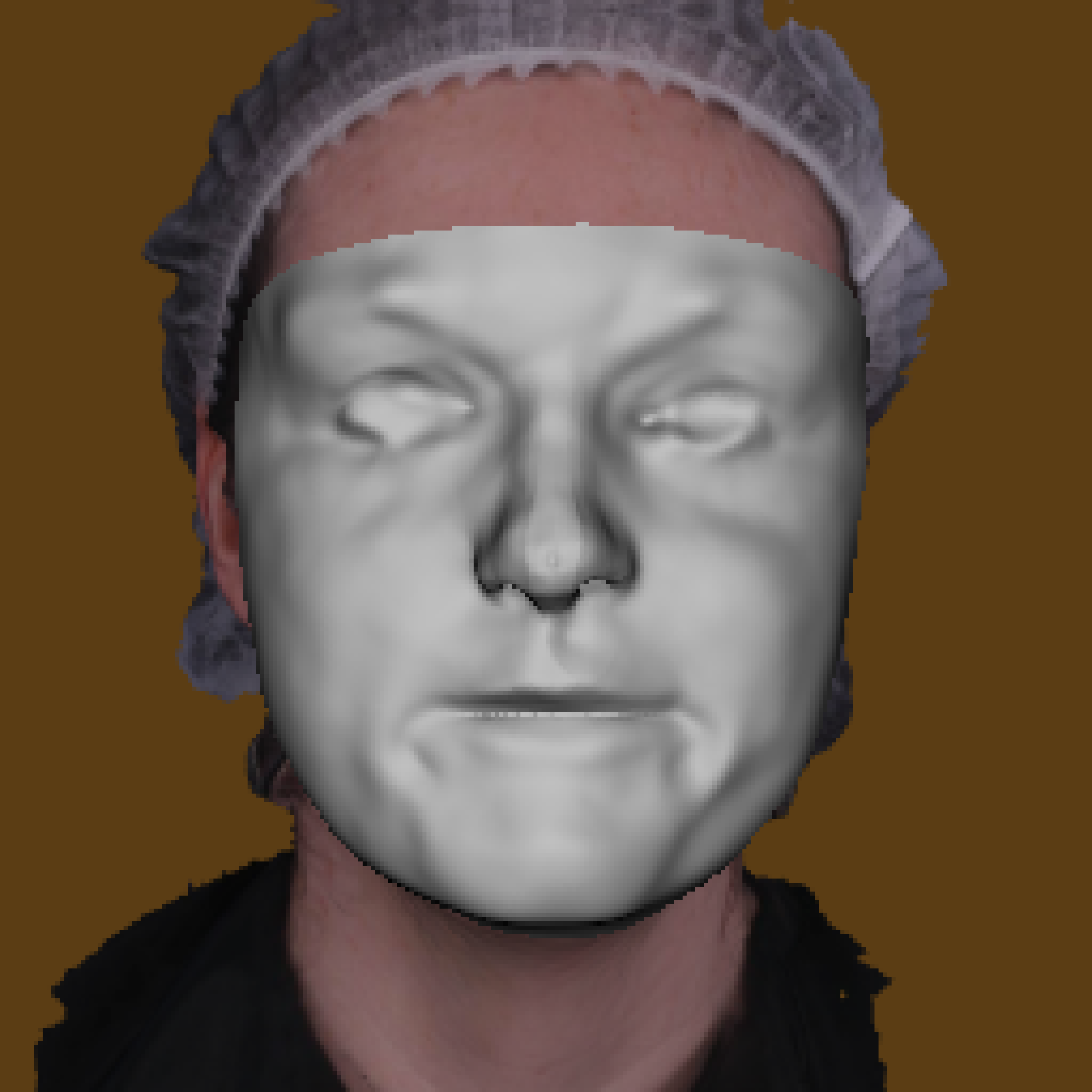}

\includegraphics[width=\galleryfigurewidth\linewidth]{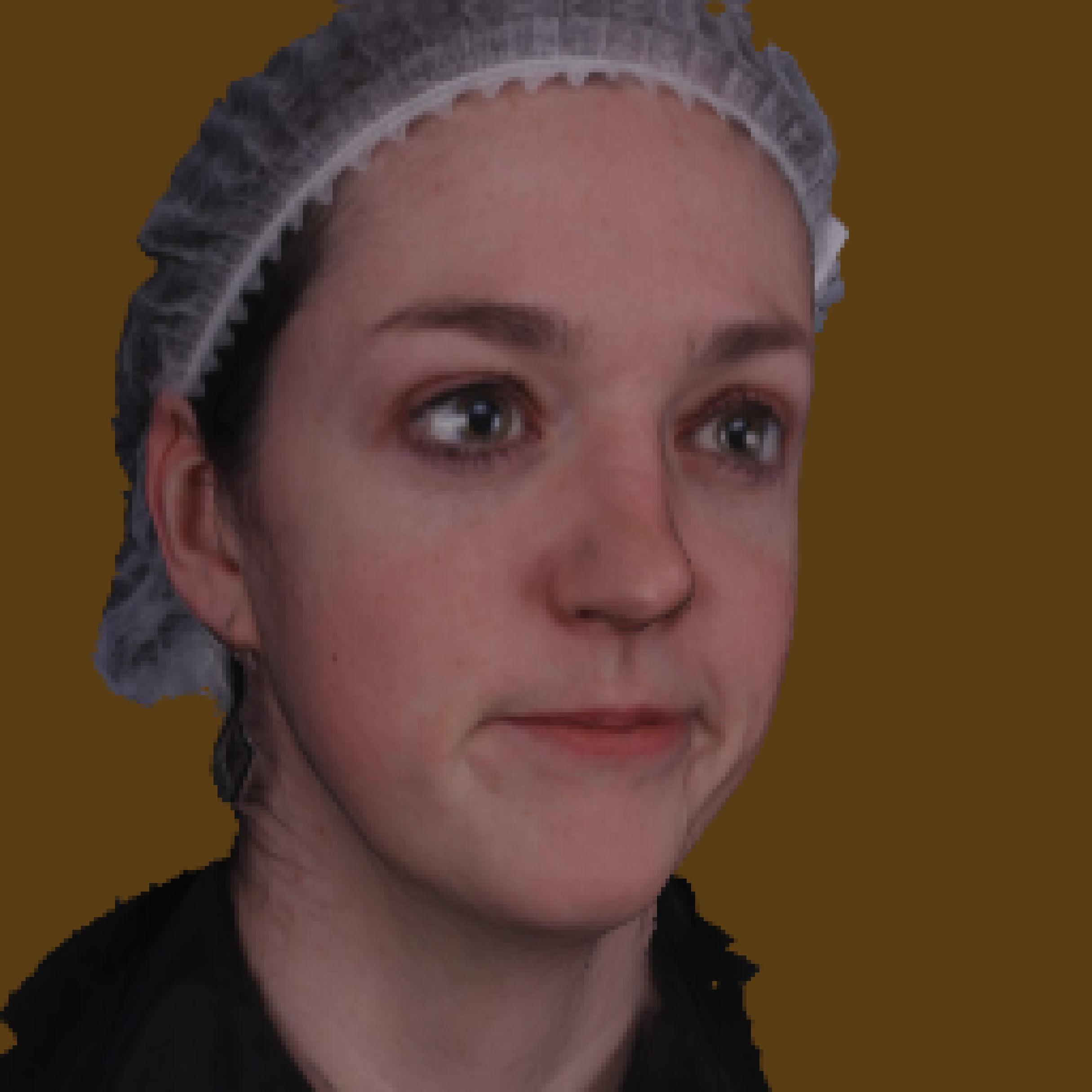}
\includegraphics[width=\galleryfigurewidth\linewidth]{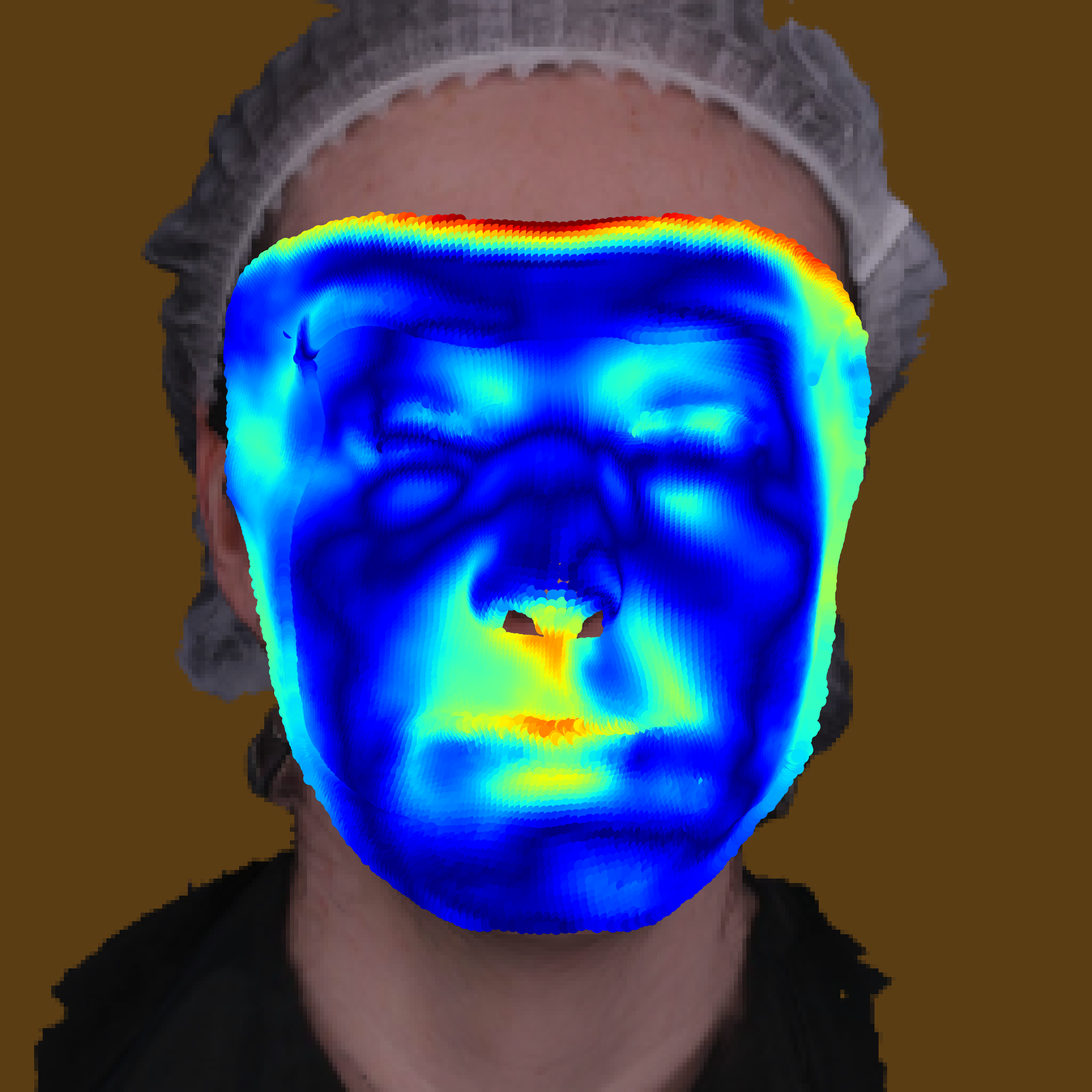}
\includegraphics[width=\galleryfigurewidth\linewidth]{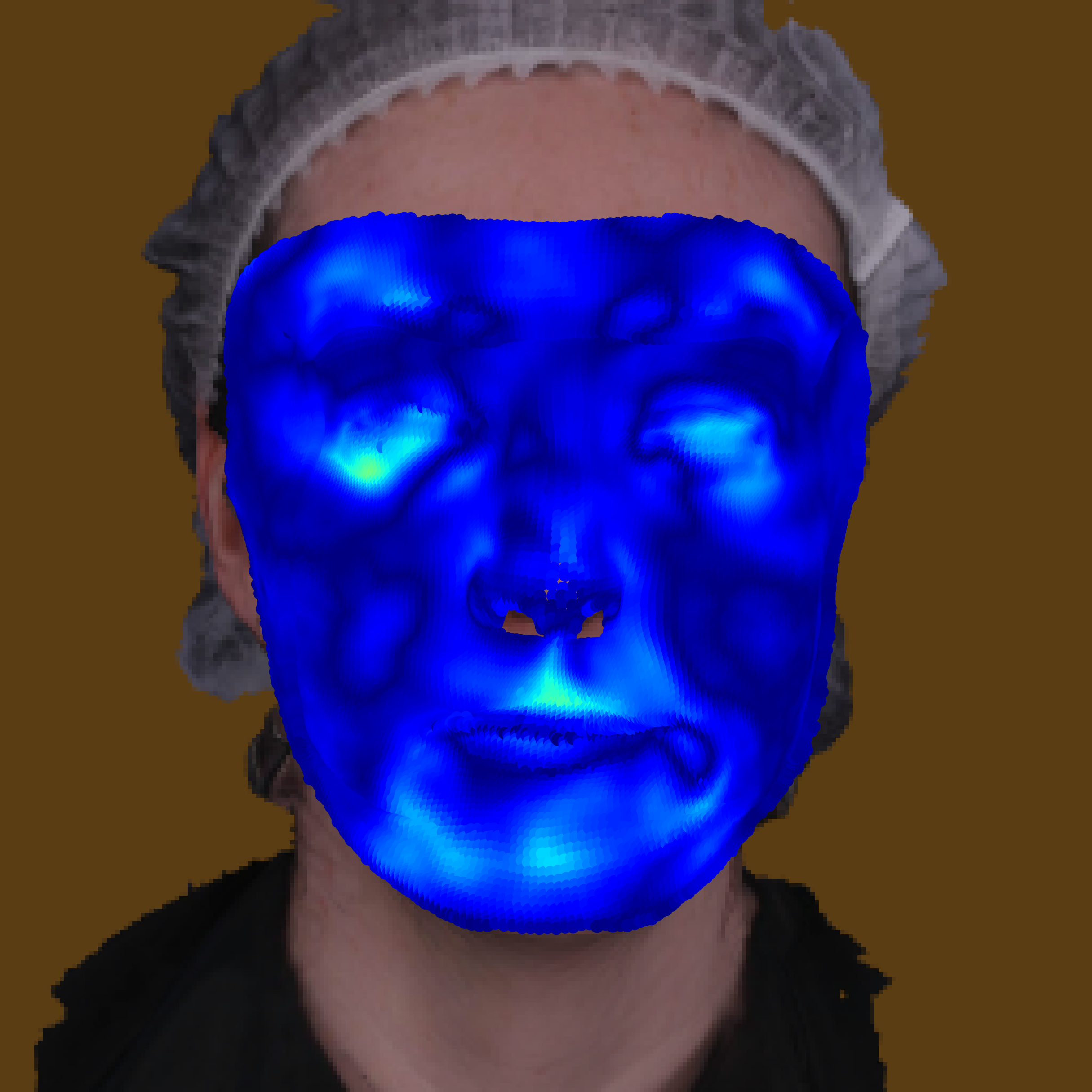}
\hspace{-15pt}
\begin{minipage}{8pt}
    \vspace{-25pt}
    \centering
    \includegraphics[width=\linewidth]{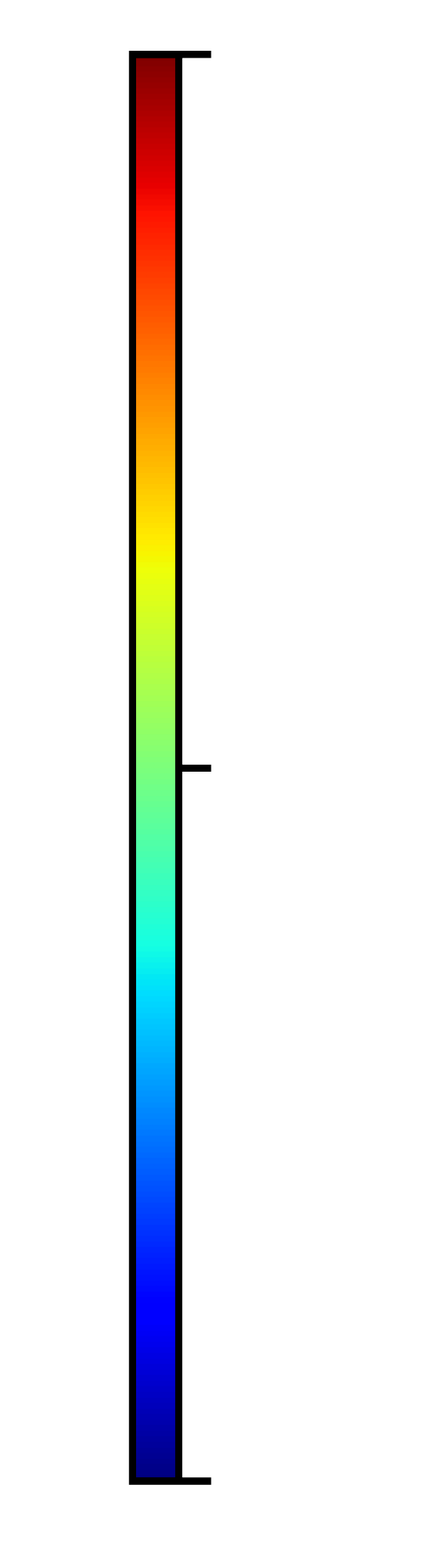}
\end{minipage}

\begin{minipage}{\galleryfigurewidth\linewidth}
    \centering
    \figtext{Input}
\end{minipage}
\begin{minipage}{\galleryfigurewidth\linewidth}
    \centering
    \figtext{DFNRMVS}
\end{minipage}
\begin{minipage}{\galleryfigurewidth\linewidth}
    \centering
    \figtext{Ours}
\end{minipage}
\end{minipage}}
\subfloat[Instance 2]{\begin{minipage}{0.495\linewidth}
\includegraphics[width=\galleryfigurewidth\linewidth]{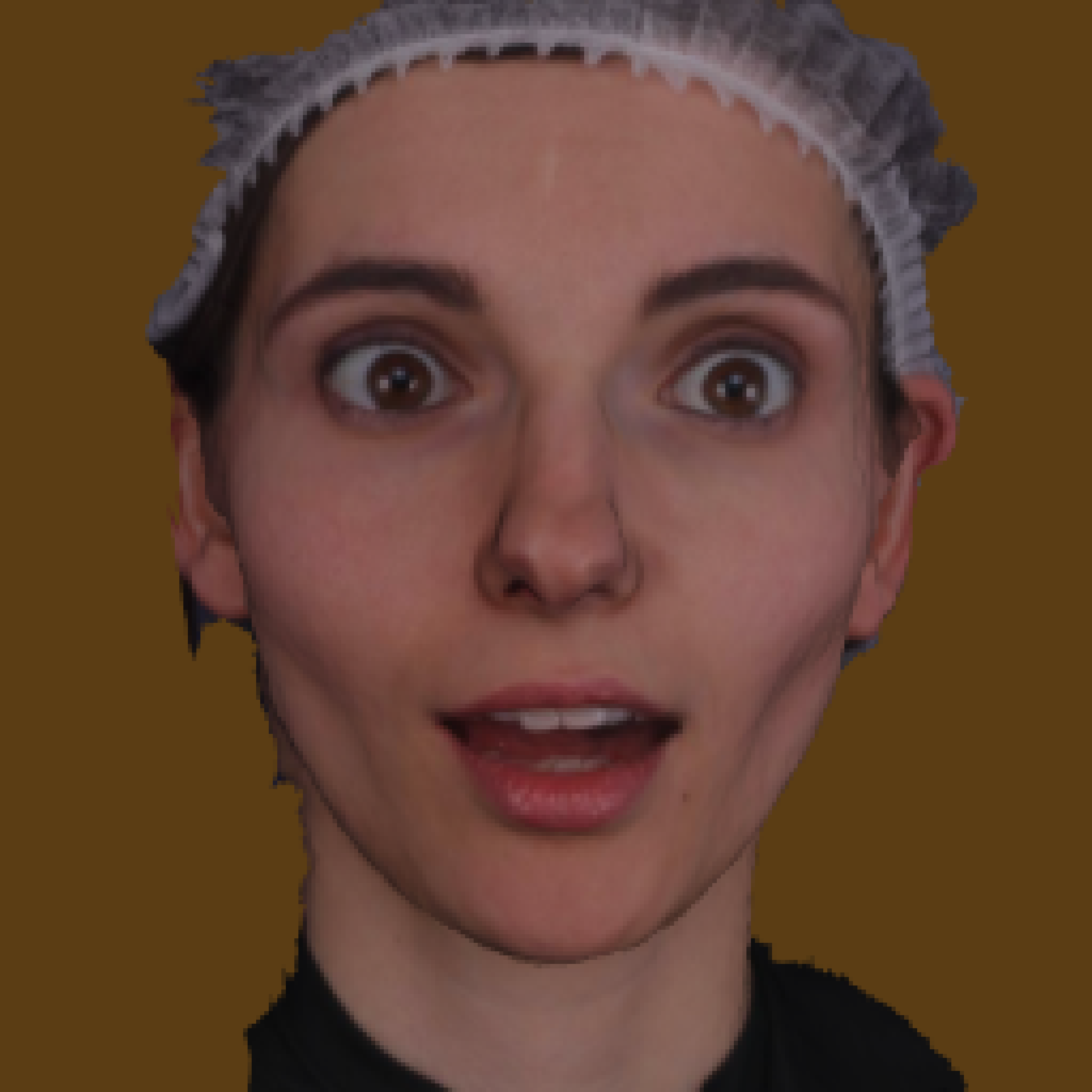}
\includegraphics[width=\galleryfigurewidth\linewidth]{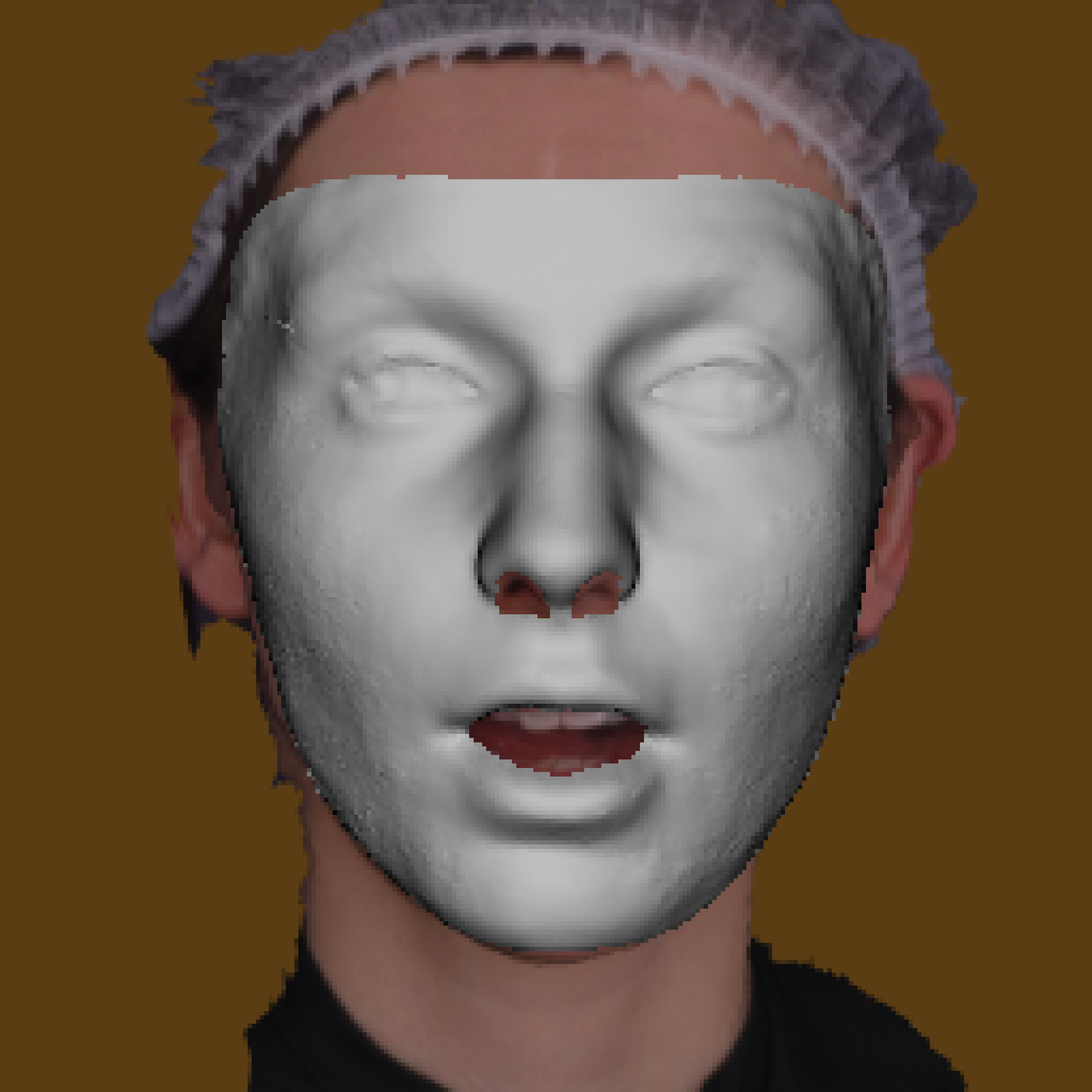}
\includegraphics[width=\galleryfigurewidth\linewidth]{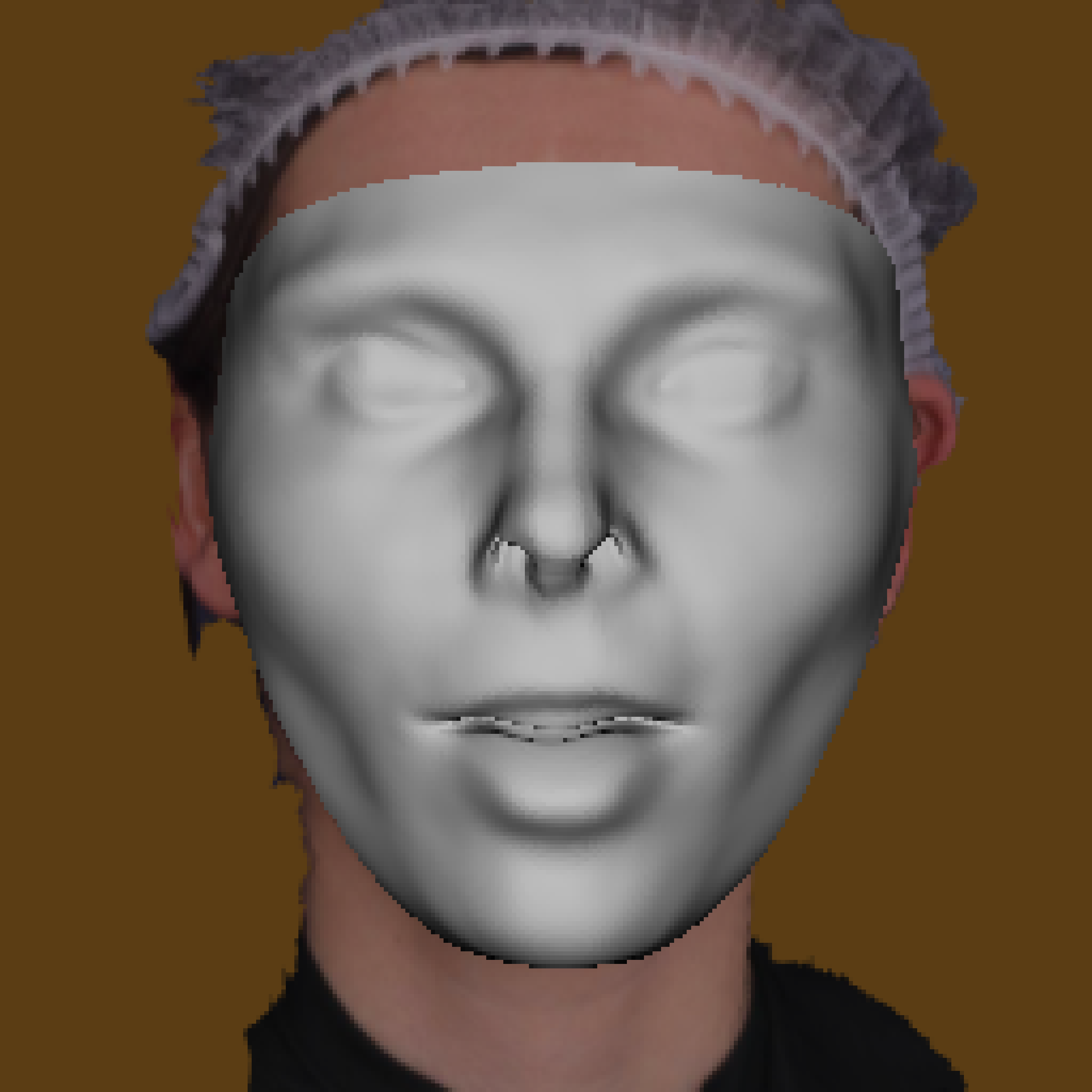}

\includegraphics[width=\galleryfigurewidth\linewidth]{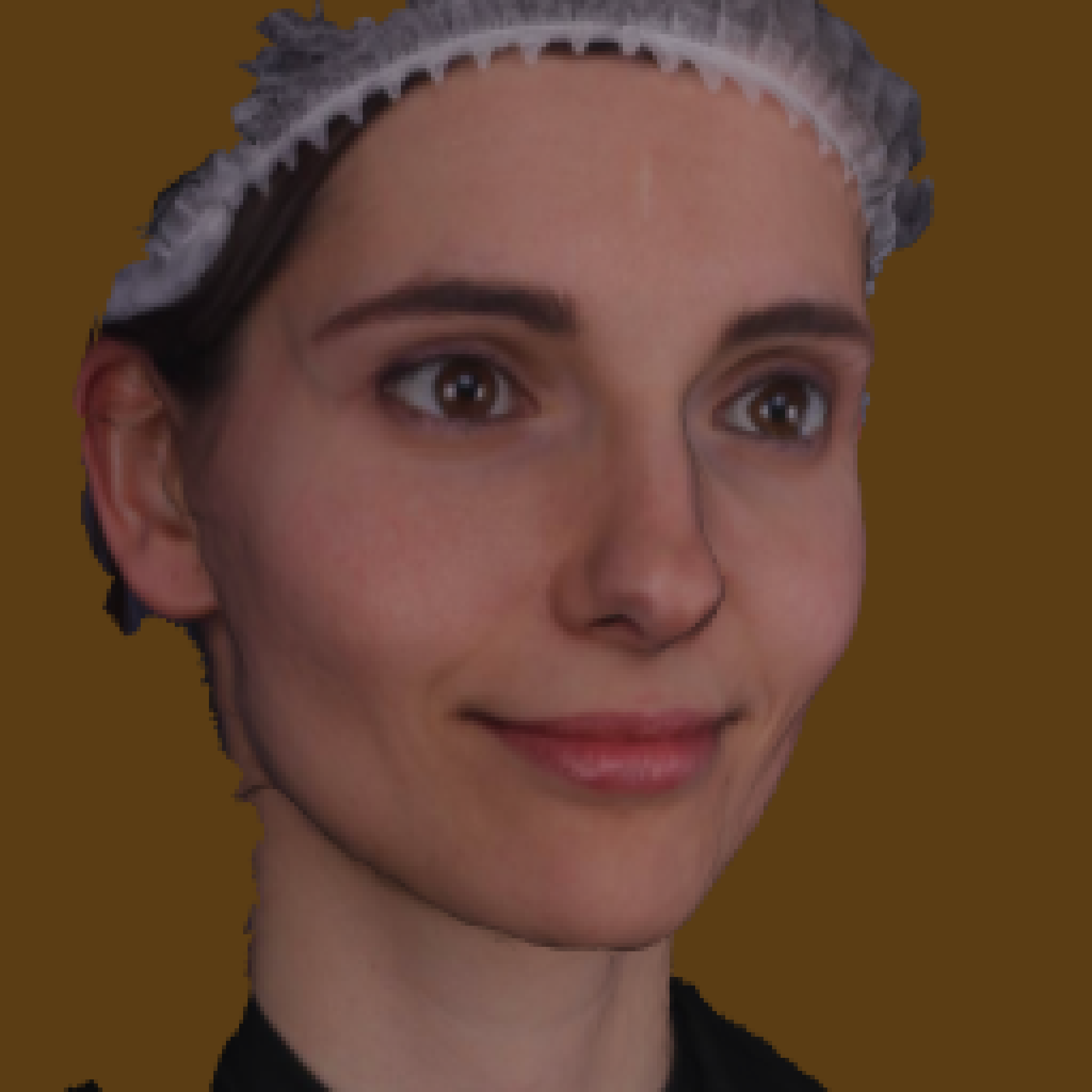}
\includegraphics[width=\galleryfigurewidth\linewidth]{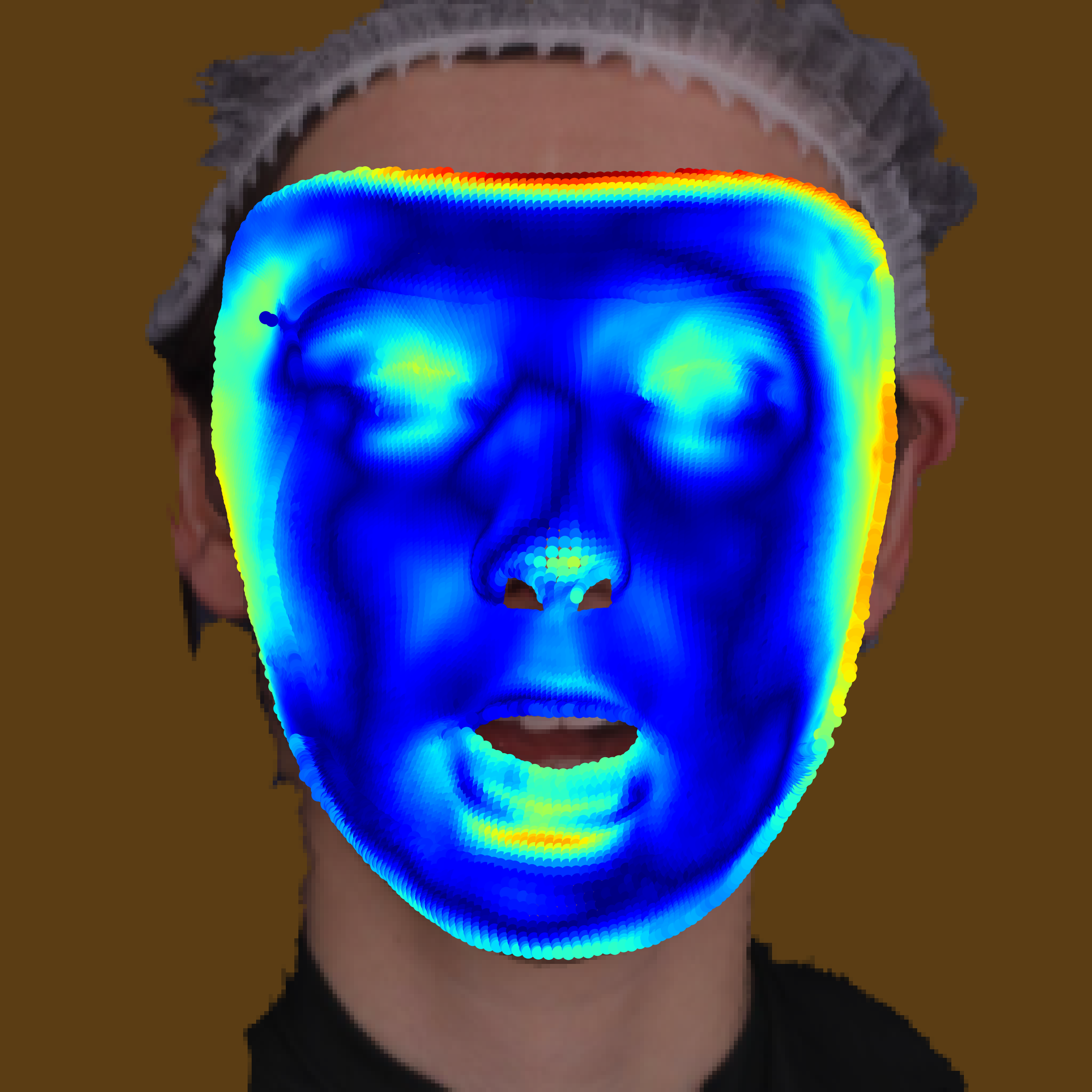}
\includegraphics[width=\galleryfigurewidth\linewidth]{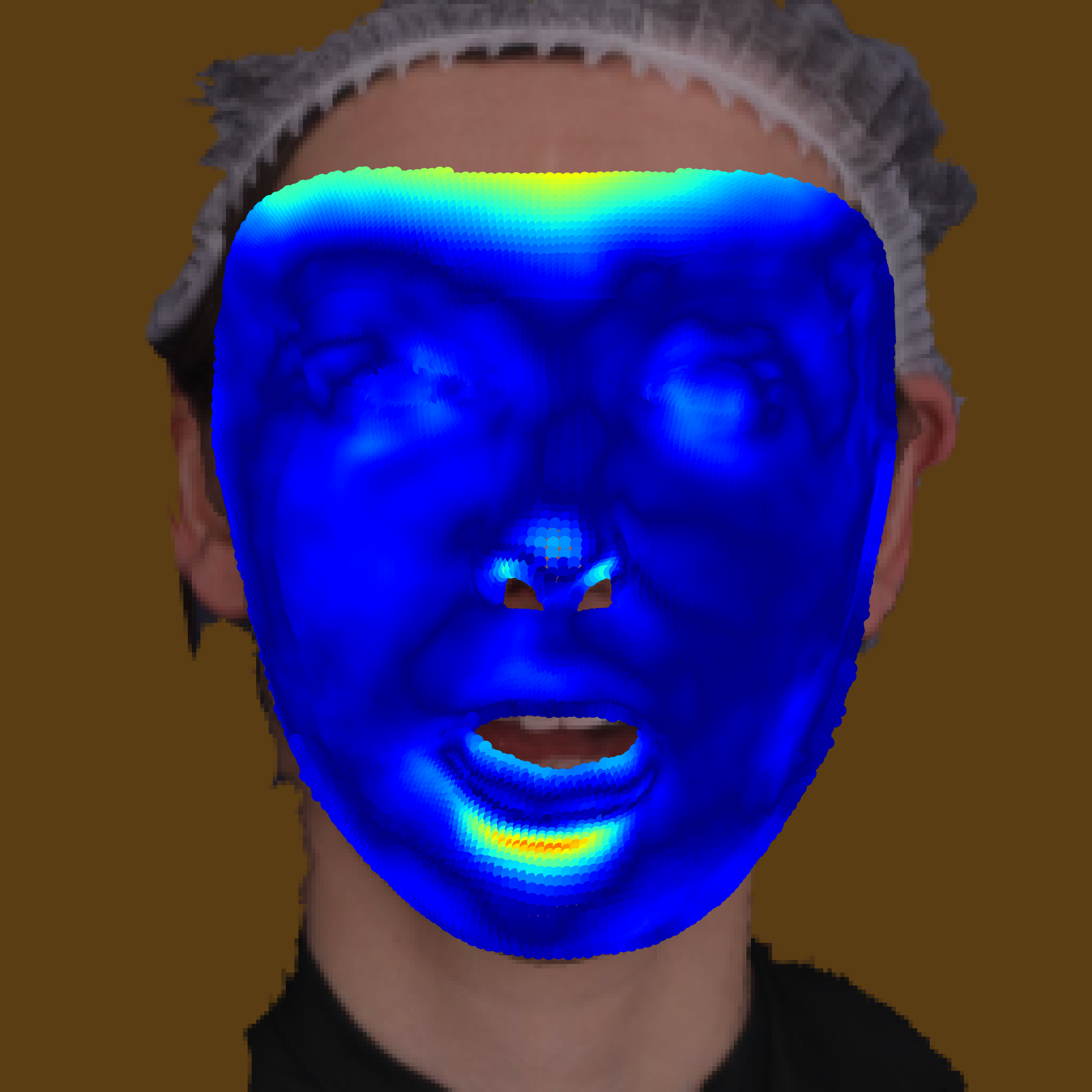}
\hspace{-15pt}
\begin{minipage}{8pt}
    \vspace{-25pt}
    \centering
    \includegraphics[width=\linewidth]{figures/gallery_stiresrc/s19/colorbar_cw_max12.png}
\end{minipage}

\begin{minipage}{\galleryfigurewidth\linewidth}
    \centering
    \figtext{Input}
\end{minipage}
\begin{minipage}{\galleryfigurewidth\linewidth}
    \centering
    \figtext{DFNRMVS}
\end{minipage}
\begin{minipage}{\galleryfigurewidth\linewidth}
    \centering
    \figtext{Ours}
\end{minipage}
\end{minipage}}
\vspace{-3pt}
%\caption{Geometry error maps on the Stirling$\slash$ESRC test set for different methods (\ie, DFNRMVS \cite{bai20dfnrmvs} and ours). The inputs are two-view images of an identity with two different expressions. In each triplet, the first column is the one of the two-view inputs, the second column is the geometry error map of DFNRMVS \cite{bai20dfnrmvs}, and the third column is our geometry error map.}
\caption{Qualitative comparison between DFNRMVS~\cite{bai20dfnrmvs} and our method on the Stirling$\slash$ESRC test set. For each instance, from left to right, we show the two input views, the results of DFNRMVS~\cite{bai20dfnrmvs}, and the ones obtained via our method, respectively. For both methods, we show the reconstructed mesh in the first row and the corresponding error map in the second row. All the results are overlaid with the first input view. The unit of the color bar is millimeter.}
\label{fig:result_gallery_stiresrc}
\label{fig:colorbar_stir}
\end{figure*}

\paragraph{Network optimization.}
The optimization of the \geonet $\geonetwork$ contains two steps. First, we optimize the \refnet to represent the surface of one scan using the 3D data (\ie, the on-surface points and the corresponding normals of this scan).
Then, the 3D training data are used to train the \deformnet $\deformnetwork$ and finetune the \refnet $\refnetwork$.
In the first step, we use the Adam optimizer with a learning rate of 1e-3.
For the second step, the learning rate is 1e-4 with a mini-batch size of 32.

The learning rate for the \rendernet optimization is set to 1e-4 with a mini-batch size of 32.
\revision{At the test time, the input images are used as the supervision to find the corresponding latent codes. Since our method does not disentangle the lighting and skin colors, we keep the \geonet fixed and finetune the \rendernet based on Eq. \eqref{eqn:loss_infer_wsw} with a small learning rate of 1e-4 to improve the representation capability for in-the-wild RGB images and reduce the domain gap.
}
The hyperparameters of Eq. \eqref{eqn:loss_infer_wsw} are similar to that of the \rendernet optimization with $\mu=0.1$ which is reduced to zero after 10 iterations.

\paragraph{Network architecture.}
We use fully connected (FC) layers with a width of 512 for our implicit neural networks.
The numbers of FC layers for \refnet, \deformnet, and \rendernet are 8, 8, and 4, respectively. 
The detailed network architecture is shown in Fig. \ref{fig:supp_netarchi}.

Note that, following \cite{idr}, we present the intersection of the ray and the surface in a differentiable way as defined in Eq. \eqref{eq:supp_onsurface}, to compute the on-surface point $\onsurface$ based on a differentiable function of the view direction $\viewdirection$ and the implicit face geometry $\network$:
\begin{equation}
\onsurface = \cameracenter + \raystep_0 \viewdirection - \frac{\viewdirection}{\nabla{\geonetwork(\onsurface_0, \geometrycode)} \viewdirection_0}{\geonetwork(\cameracenter + \raystep_0 \viewdirection, \geometrycode)}
\label{eq:supp_onsurface}
\end{equation}
where $\cameracenter + \raystep_0 \viewdirection$ is the first intersection found by the differentiable sphere-tracing algorithm, and $\raystep_0$ is the initial ray step. Besides,  $\onsurface_0 = \cameracenter + \raystep_0 \viewdirection$ is the initial intersection position. $\viewdirection_0$ is the initial value of the view direction.

Following several previous methods about implicit neural representations \cite{idr,mildenhall2020nerf_sdf,yenamandra2020i3dmm}, we use the Fourier positional encoding \cite{tancik2020fourier} for the input 3D coordinates of the \refnet and the \deformnet to reduce the difficulty of learning high-frequency functions.
For each of the three coordinates (\ie, $x_i$) of $\onsurface=\{x_i\}^3_{i=1}$, the positional encoding is $w^{K}(x_i) = \sum^K_{k=0} (\text{cos}(2^{k} \pi x_i) + \text{sin}(2^{k} \pi x_i))$, where $K=6$. For the view direction $\viewdirection$, we also use such a positional encoding as $w^{4}(\cdot)$.

\paragraph{Timing statistics.}
We implement our method using Pytorch with NVIDIA 2080Ti GPUs. The training time is about two days with 8 GPUs. The test-time reconstruction for one instance with 2$\sim$4 input views is about two hours with one GPU.

\subsection{Experimental Setup}

\paragraph{Datasets.}
We conduct experiments on two benchmarks for 3D face reconstruction as listed below.
\begin{itemize}%[leftmargin=*]
\item The Stirling{\slash}ESRC dataset~\cite{stir3d} provides more than 1K high-quality 3D scans and is built upon more than 130 subjects with 8 different expressions.
For each scan, a pair of RGB images taken from yaw angles of $\pm 45^{\circ}$ are used as the texture.
We split this dataset into training and testing sets containing $95$ and $35$ subjects, respectively, in the same way as \cite{bai20dfnrmvs}. % for fair comparison.

\item The Bosphorus dataset~\cite{savran2008bosphorus} contains $106$ subjects with $35$ expressions and $13$ poses.
For each subject, the images of non-neutral expressions are under the frontal view, while those of the neutral expression are under various poses.
Following \cite{deng2019accurate_weak-super_mv,bai20dfnrmvs,bai2021riggable}, we adopt this dataset to evaluate the performance of our method under a two-view setting.
Specifically, we select a non-neutral frontal view image and another image of a neutral expression under a yaw angle of $-30^{\circ}$ for each instance. Hence, the overall test set contains $453$ pairs of images.
\end{itemize}

\begin{figure*}%[ht]
\centering
\subfloat[Instance 1]{\begin{minipage}{0.49\linewidth}
\includegraphics[width=\galleryfigurewidth\linewidth]{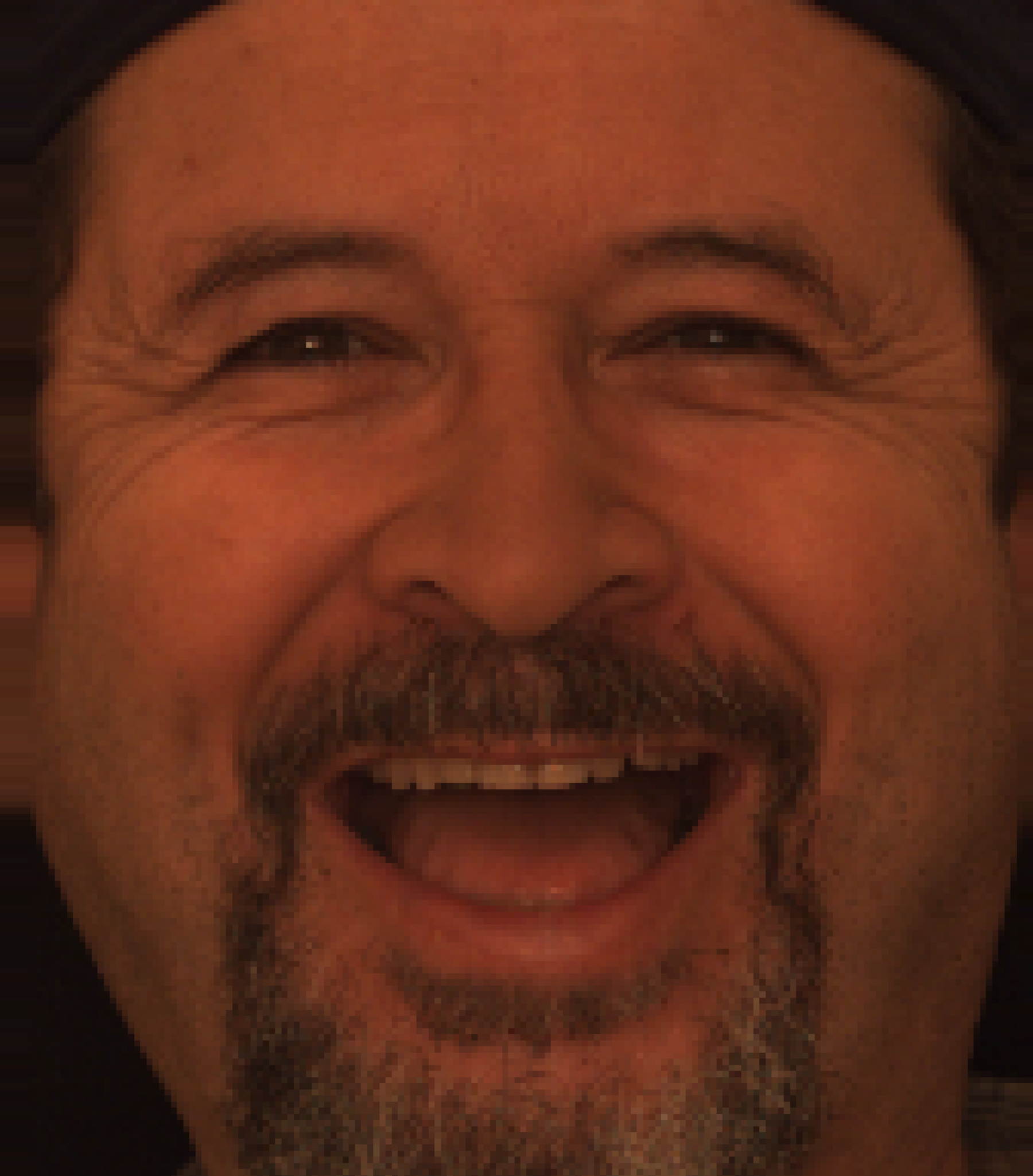}
\includegraphics[width=\galleryfigurewidth\linewidth]{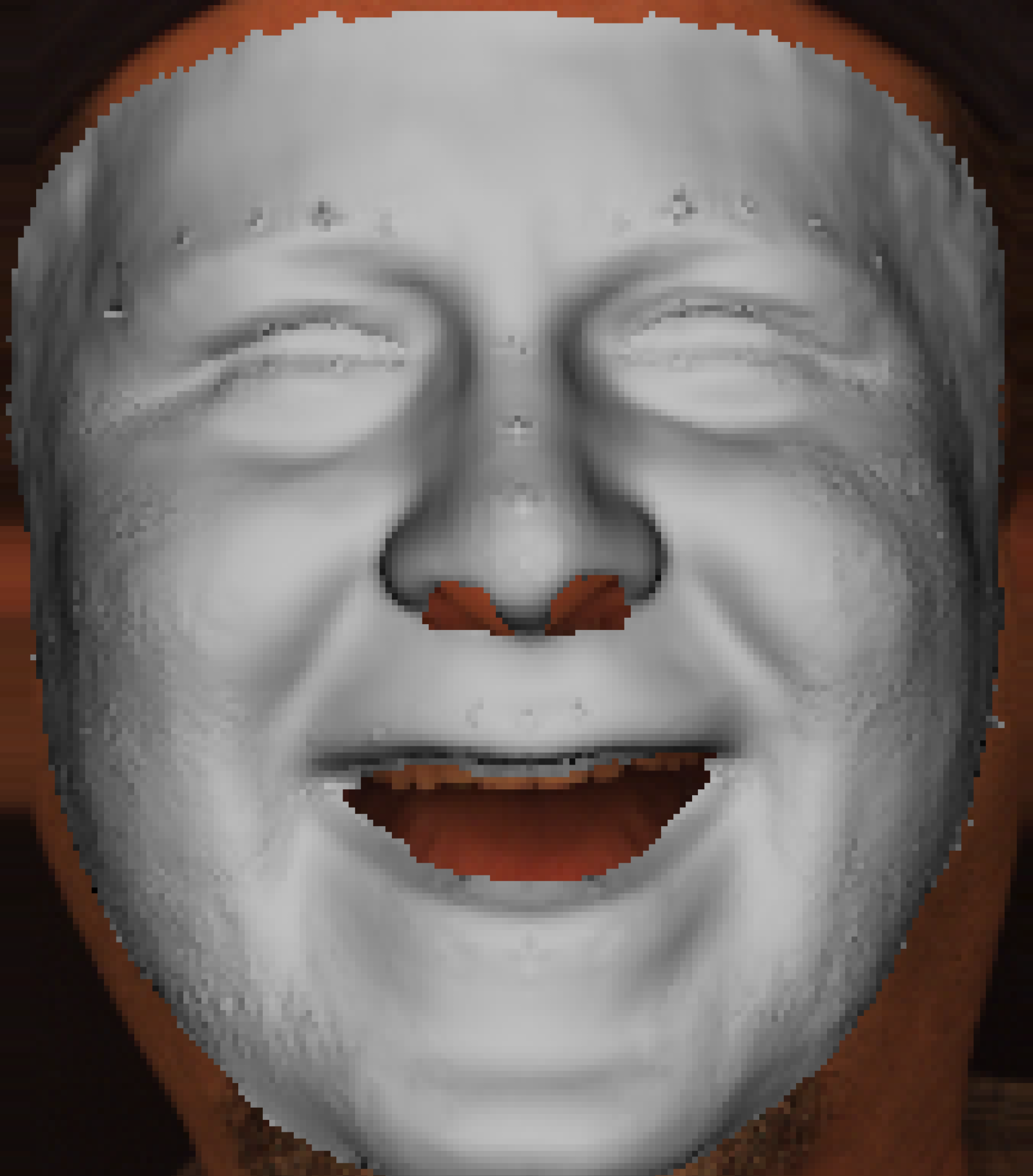}
\includegraphics[width=\galleryfigurewidth\linewidth]{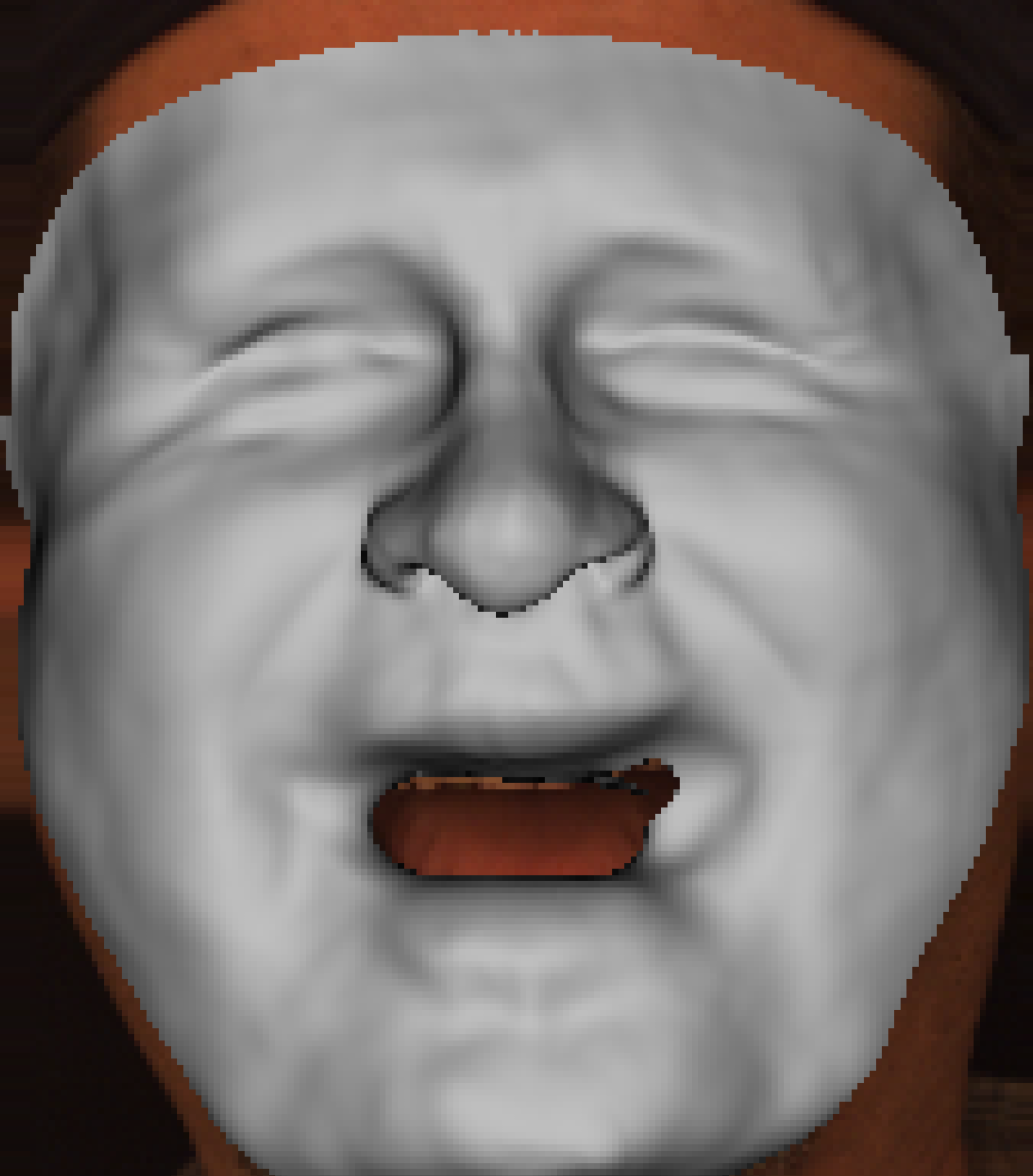}

\includegraphics[width=\galleryfigurewidth\linewidth]{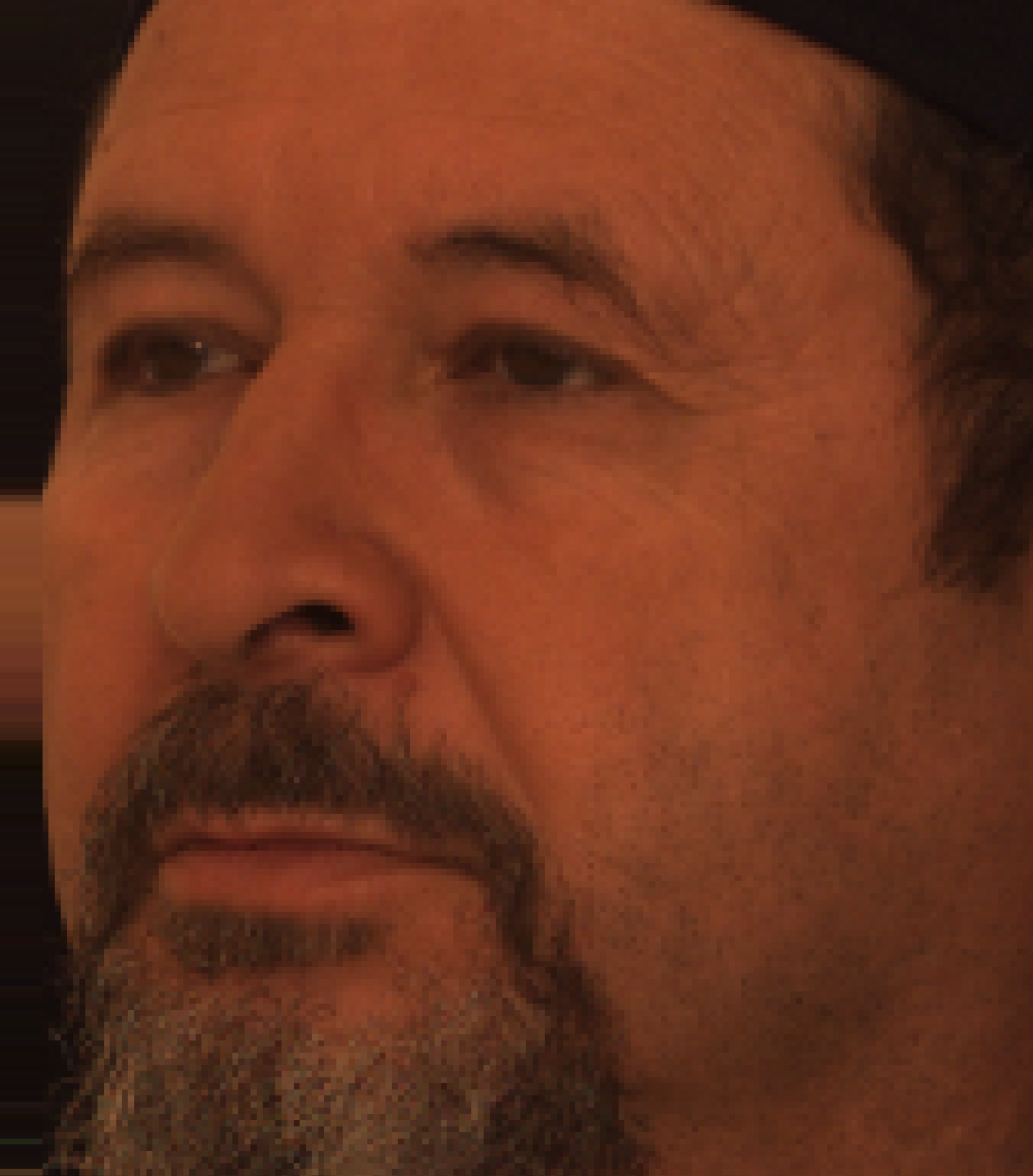}
\includegraphics[width=\galleryfigurewidth\linewidth]{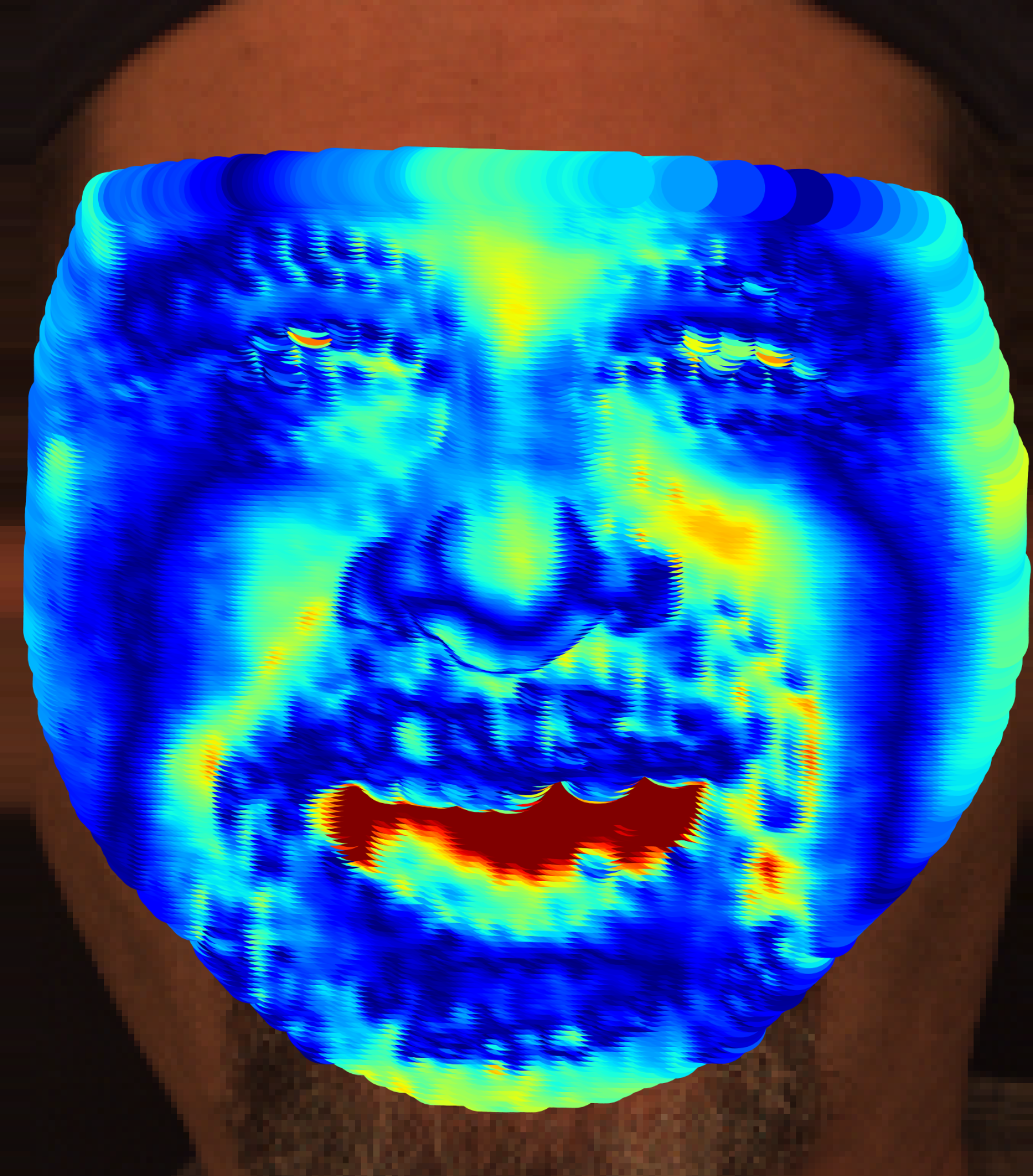}
\includegraphics[width=\galleryfigurewidth\linewidth]{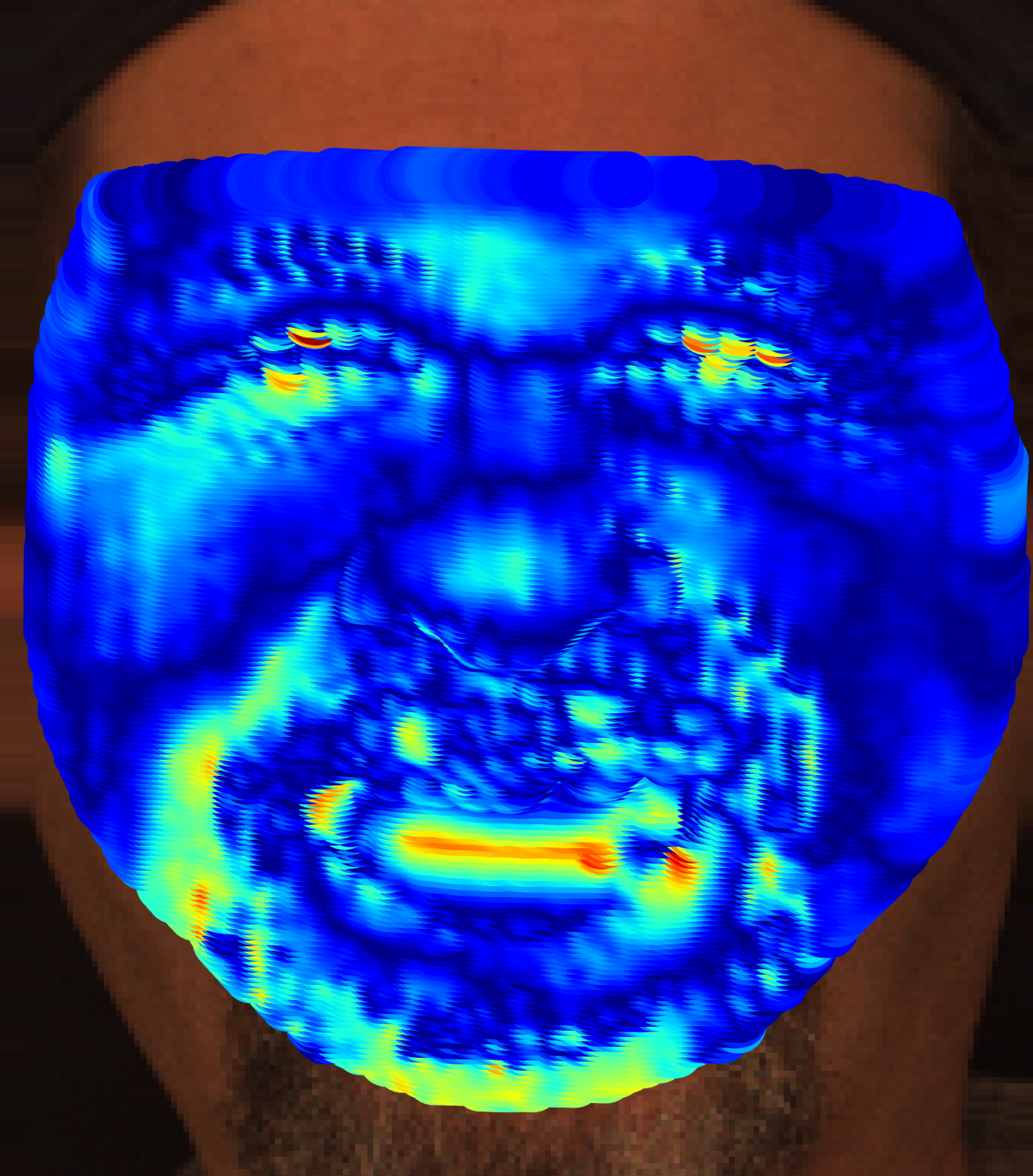}
\hspace{-13pt}
\begin{minipage}{6pt}
    \vspace{-20pt}
    \centering
    \includegraphics[width=\linewidth]{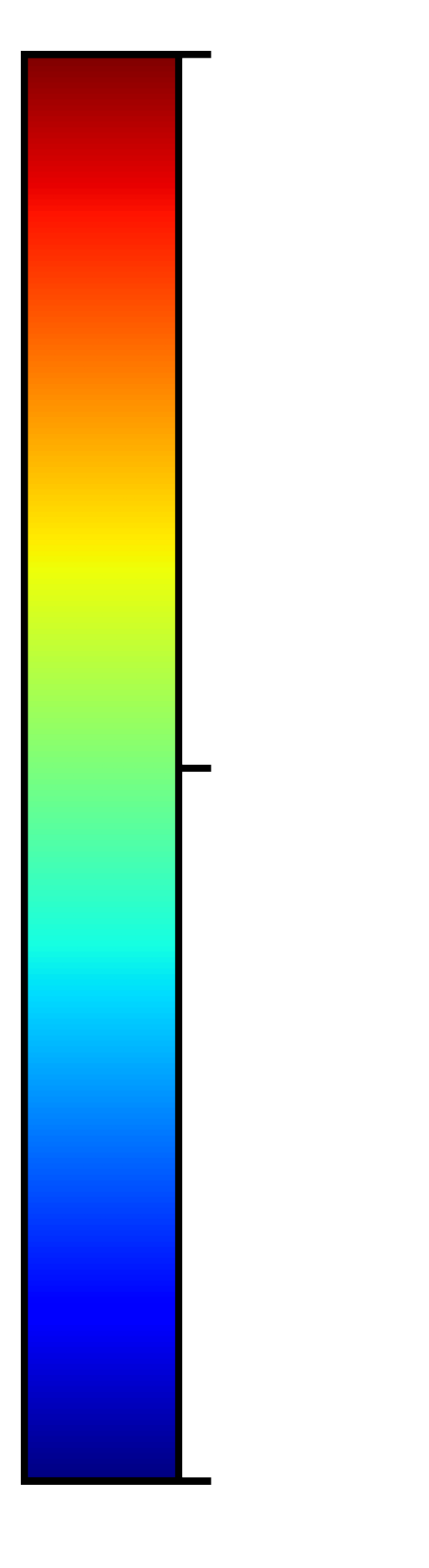}
\end{minipage}

\begin{minipage}{\galleryfigurewidth\linewidth}
    \centering
    \figtext{Input}
\end{minipage}
\begin{minipage}{\galleryfigurewidth\linewidth}
    \centering
    \figtext{DFNRMVS}
\end{minipage}
\begin{minipage}{\galleryfigurewidth\linewidth}
    \centering
    \figtext{Ours}
\end{minipage}
\end{minipage}}
\subfloat[Instance 2]{\begin{minipage}{0.49\linewidth}
\includegraphics[width=\galleryfigurewidth\linewidth]{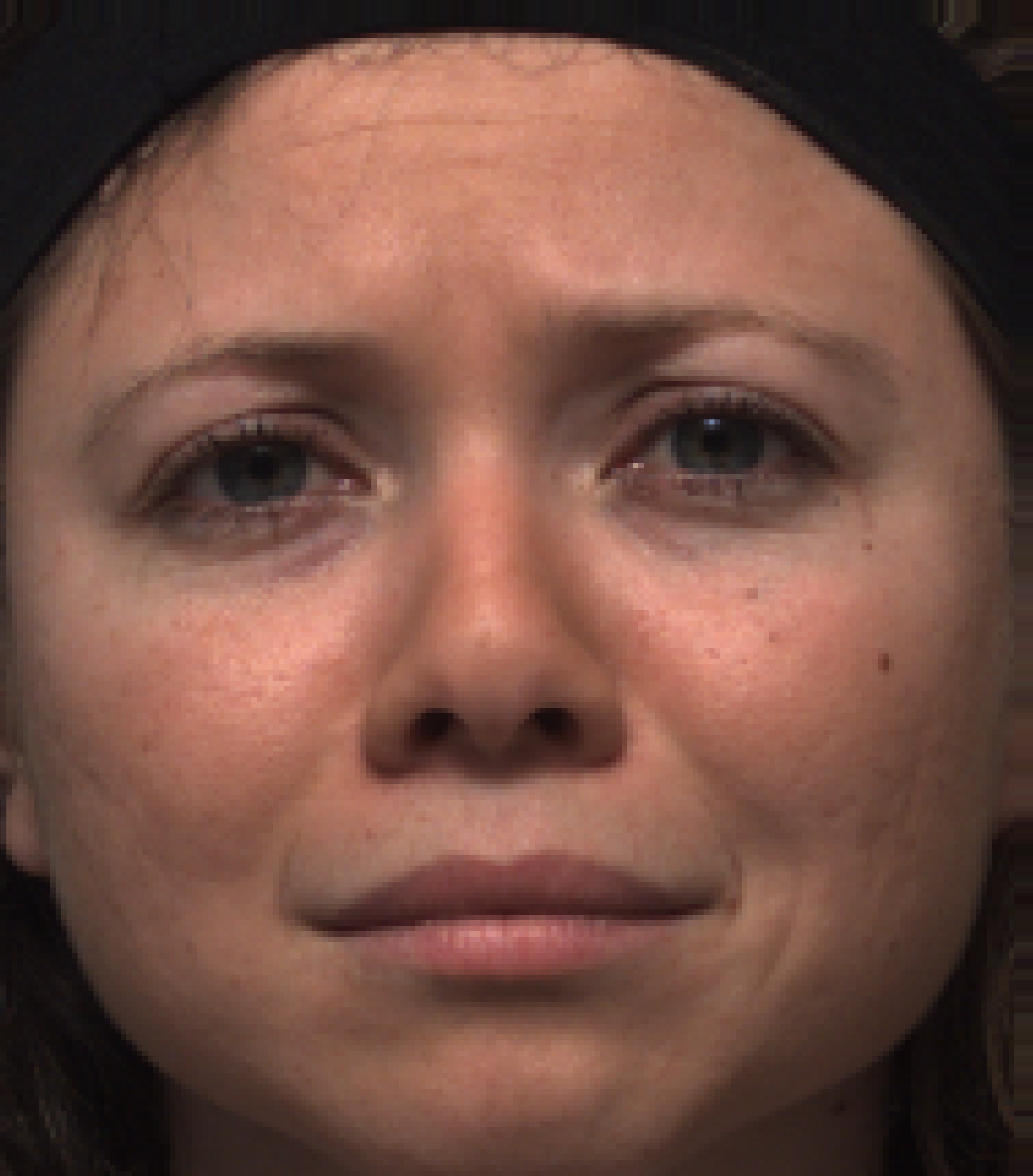}
\includegraphics[width=\galleryfigurewidth\linewidth]{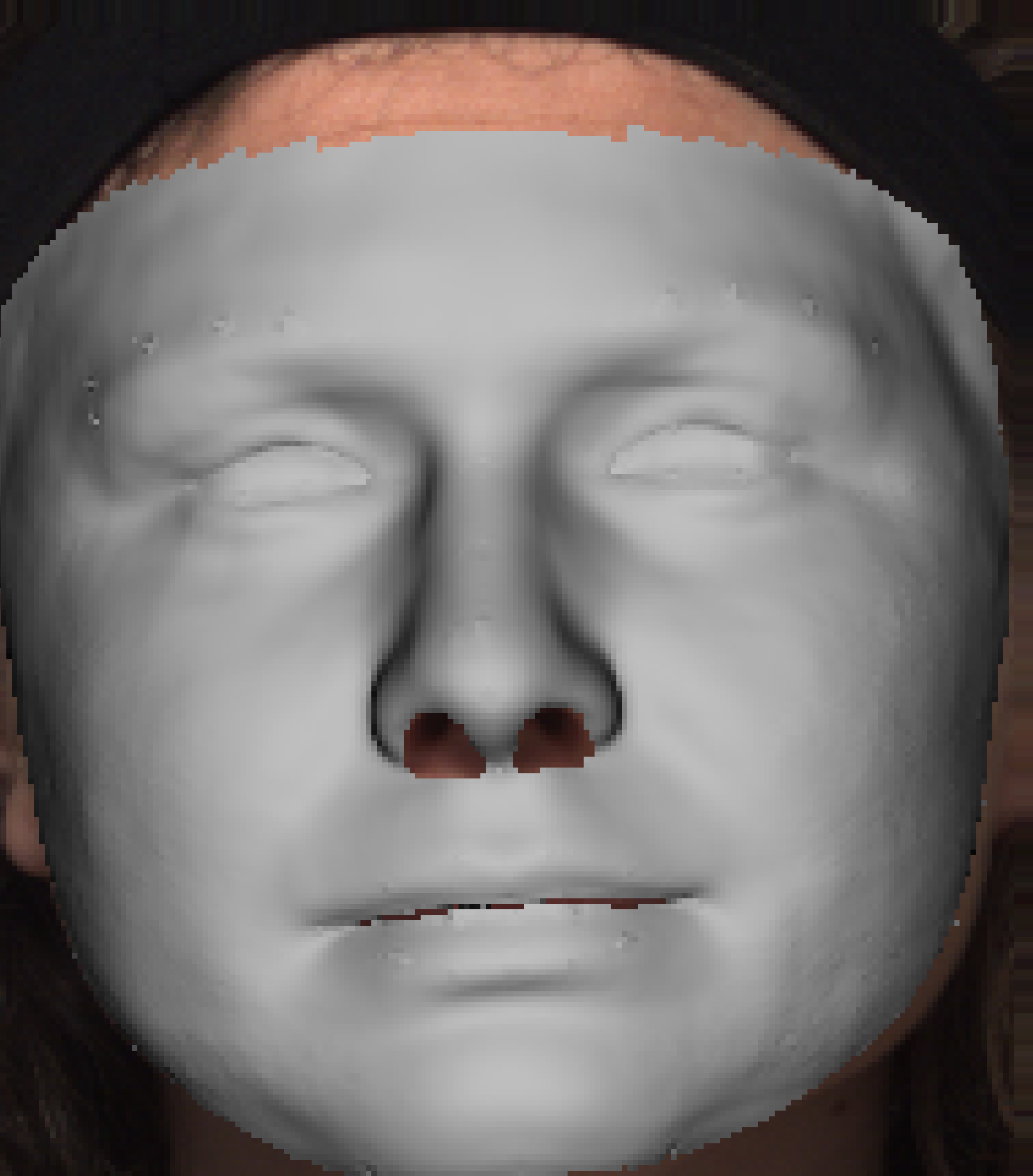}
\includegraphics[width=\galleryfigurewidth\linewidth]{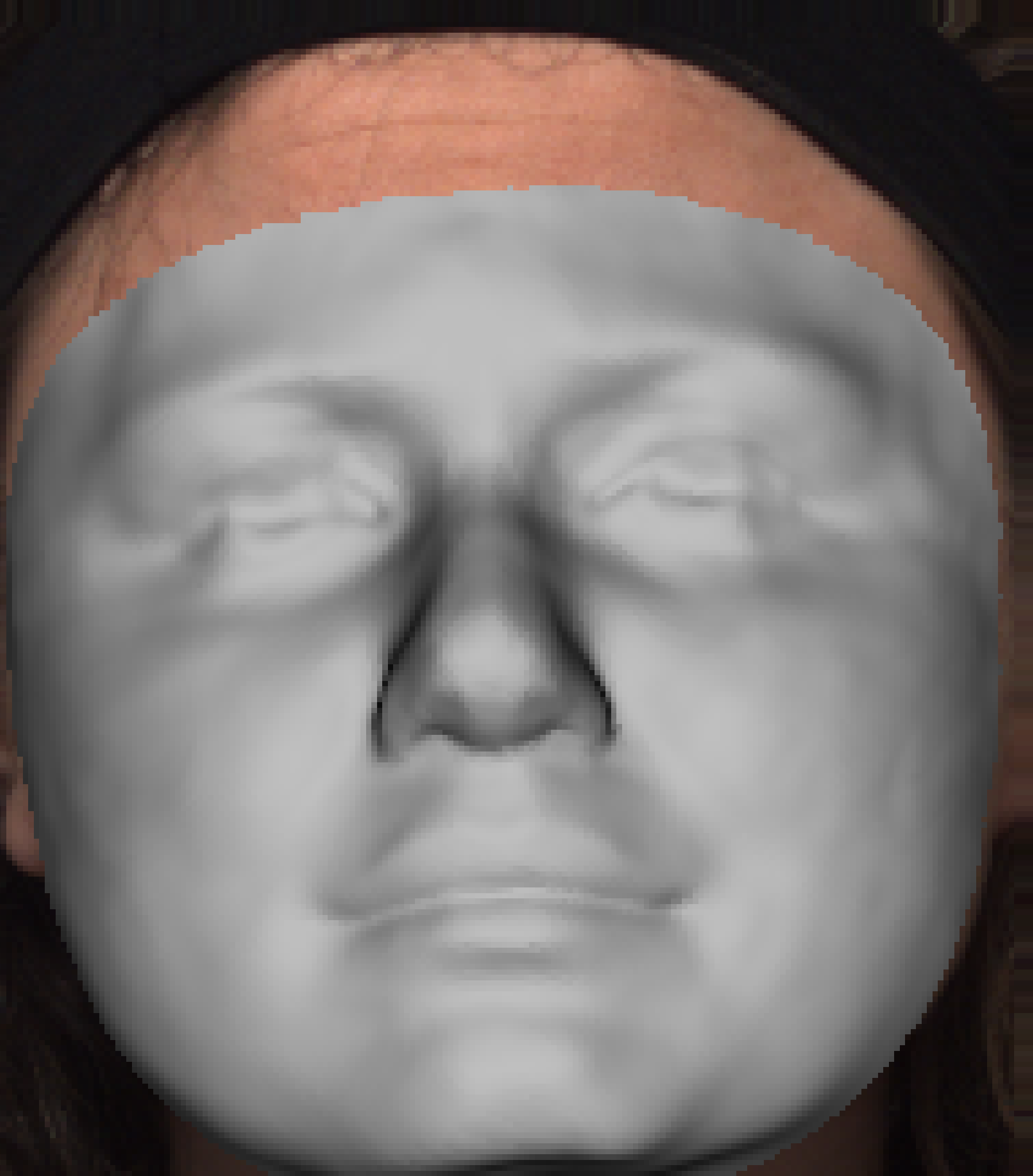}

\includegraphics[width=\galleryfigurewidth\linewidth]{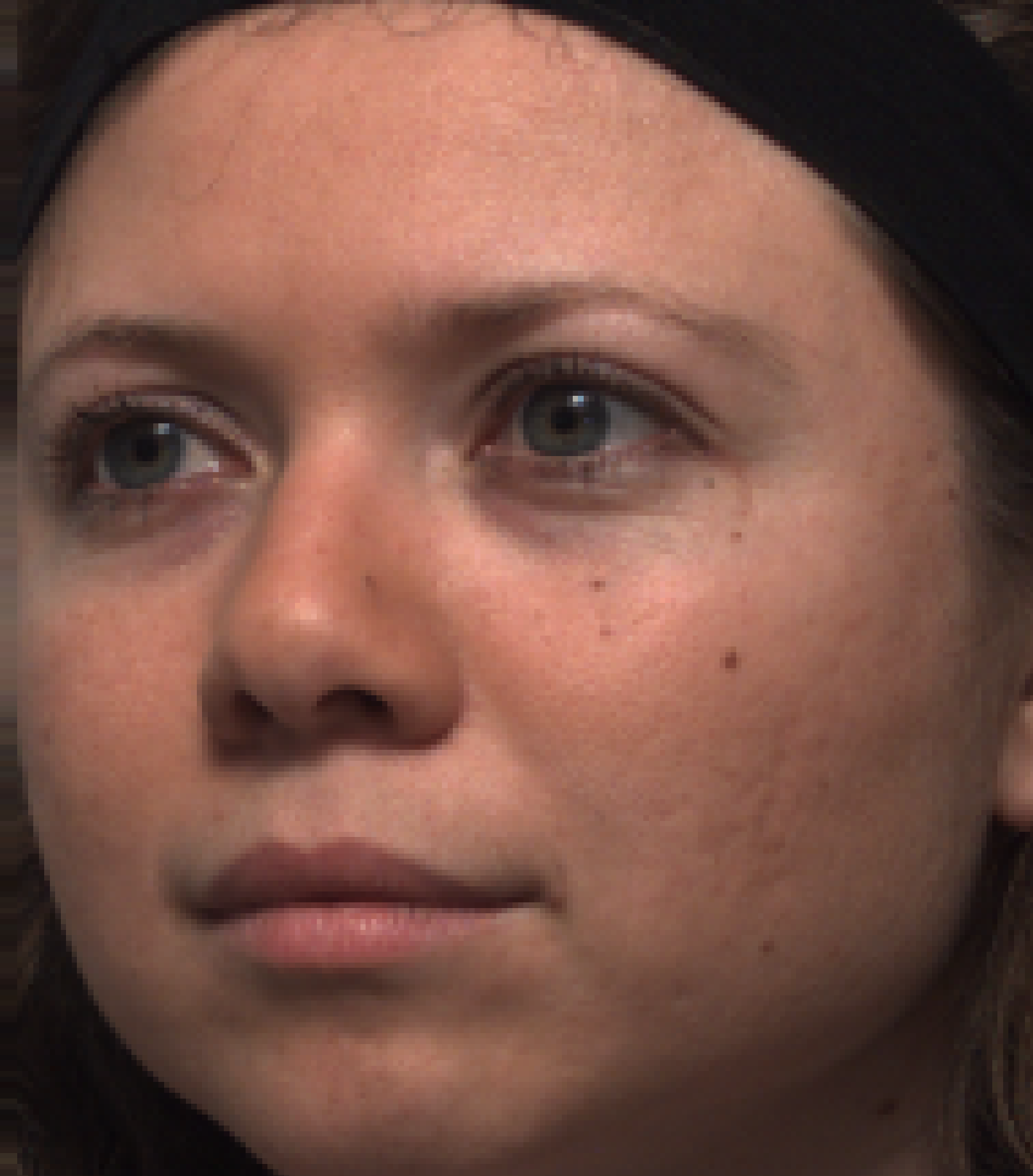}
\includegraphics[width=\galleryfigurewidth\linewidth]{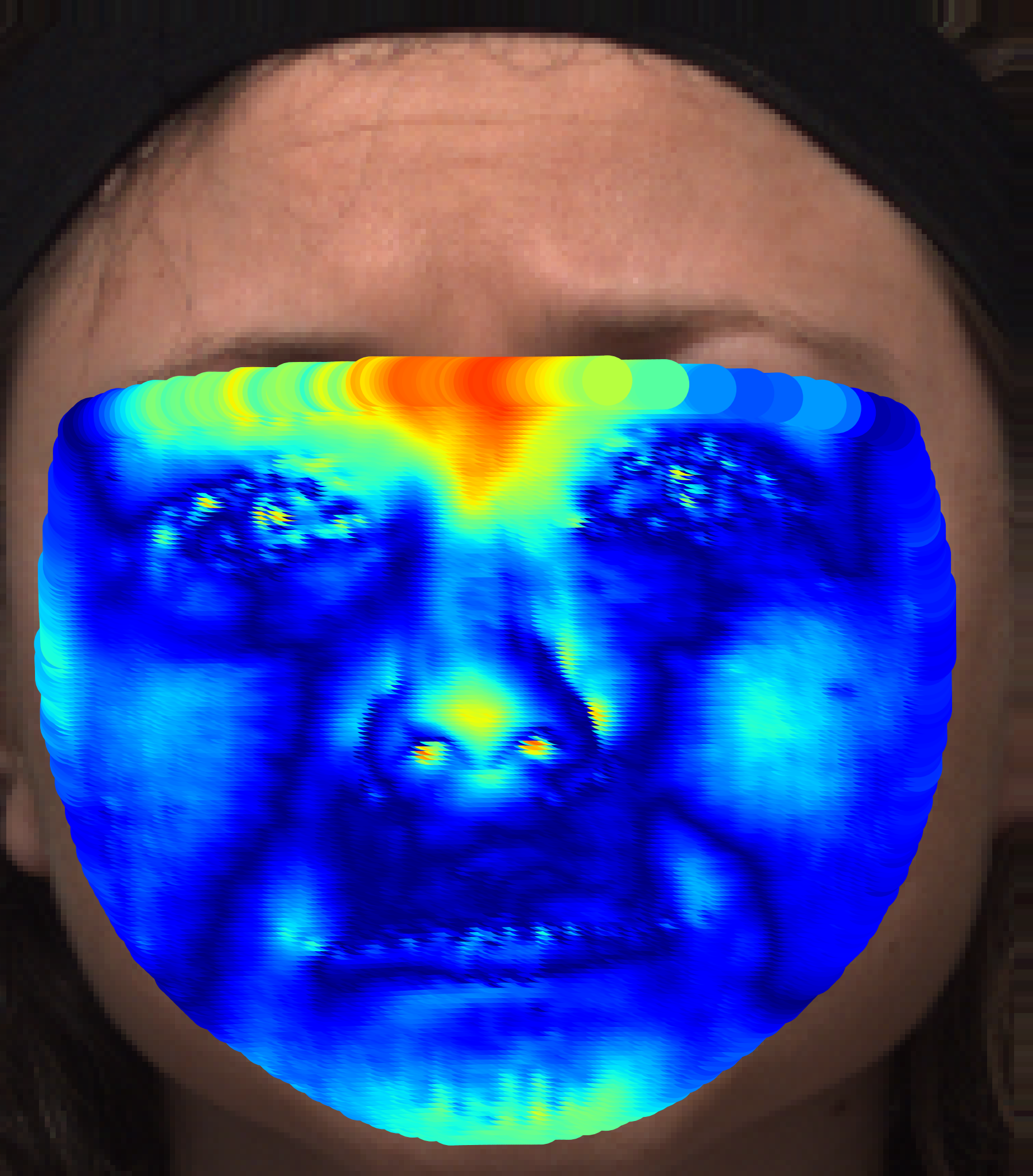}
\includegraphics[width=\galleryfigurewidth\linewidth]{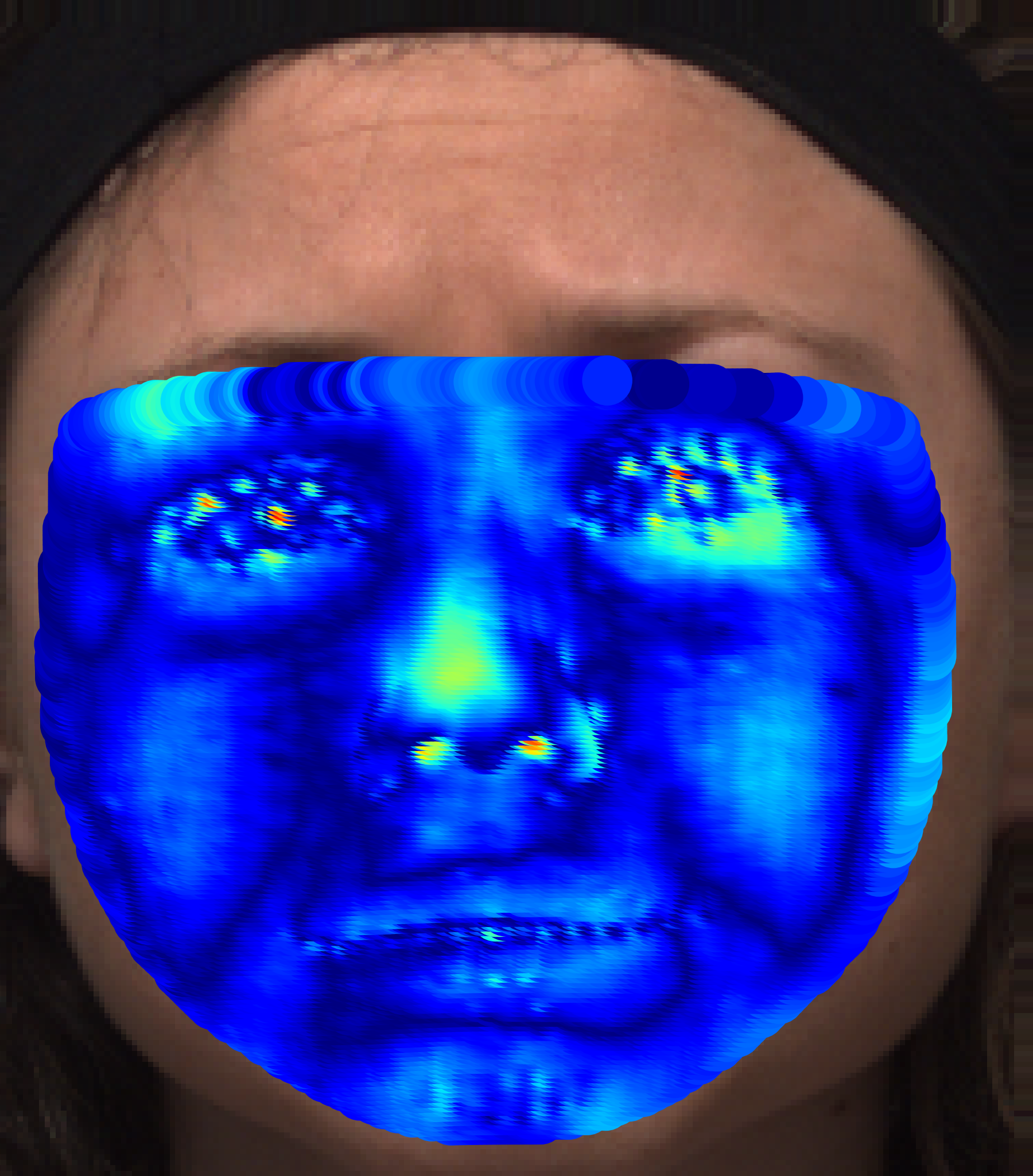}
\hspace{-13pt}
\begin{minipage}{5pt}
    \vspace{-20pt}
    \centering
    \includegraphics[width=\linewidth]{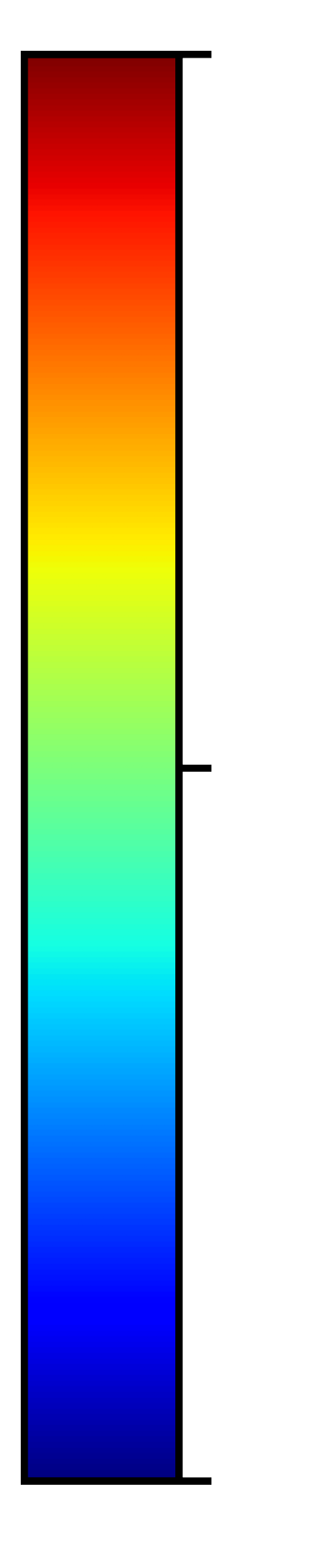}
\end{minipage}

\begin{minipage}{\galleryfigurewidth\linewidth}
    \centering
    \figtext{Input}
\end{minipage}
\begin{minipage}{\galleryfigurewidth\linewidth}
    \centering
    \figtext{DFNRMVS}
\end{minipage}
\begin{minipage}{\galleryfigurewidth\linewidth}
    \centering
    \figtext{Ours}
\end{minipage}
\end{minipage}}
\vspace{-3pt}
\caption{Qualitative comparison between DFNRMVS \cite{bai20dfnrmvs} and our method on the Bosphorus dataset~\cite{savran2008bosphorus}. The images are arranged in the same way as Figure~\ref{fig:result_gallery_stiresrc}.}
\label{fig:result_gallery_bosp}
\end{figure*}

\paragraph{Evaluation protocols.}
Following previous methods~\cite{deng2019accurate_weak-super_mv, bai20dfnrmvs,bai2021riggable}, we use the Euclidean distance between the ground truth 3D face surface points and the aligned output mesh to evaluate the geometric error.
The average geometric error (Mean, in mm) and the standard deviation (STD, in mm) among all test samples are computed and reported in our quantitative evaluations.
The alignment contains two steps: (i) we first use the ground truth landmarks (\eg, 7 landmarks provided in Bosphorus dataset \cite{savran2008bosphorus}) and the landmark points in our reconstruction results to achieve rough alignment \cite{schonemann1966generalized_procrustes_align}; (ii) we use the rigid ICP algorithm~\cite{zhou2018open3dicp} to further improve the alignment between our prediction and the ground truth.
Besides, the ground truth is cropped to reduce the noise based on the corresponding 3D landmarks.
Note that those strategies are the same as previous methods~\cite{deng2019accurate_weak-super_mv,bai20dfnrmvs,bai2021riggable}.

\subsection{Qualitative Results}
We present qualitative comparisons between DFNRMVS~\cite{bai20dfnrmvs} and our method on both Stirling{\slash}ESRC and Bosphorus datasets in Figs.~\ref{fig:colorbar_stir} and~\ref{fig:result_gallery_bosp}, respectively.
For each test instance in both figures, we show the two input views on the left, the result by DFNRMVS~\cite{bai20dfnrmvs} in the middle, and our result on the right side.
We provide the flat-shaded face mesh in the first row and the corresponding error map in the second row.

\newcommand\suppw{0.235}
\begin{figure*}
\centering
\subfloat[View 1]{\begin{minipage}{0.495\linewidth}
\includegraphics[width=\suppw\linewidth]{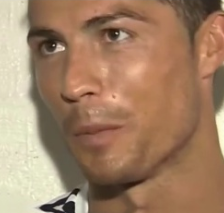}
\includegraphics[width=\suppw\linewidth]{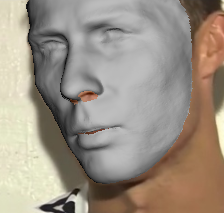}
\includegraphics[width=\suppw\linewidth]{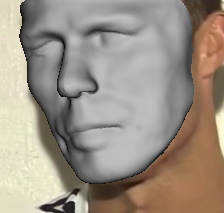}
\includegraphics[width=\suppw\linewidth]{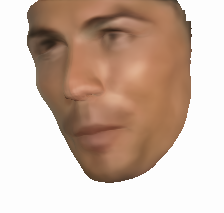}

\includegraphics[width=\suppw\linewidth]{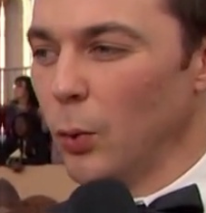}
\includegraphics[width=\suppw\linewidth]{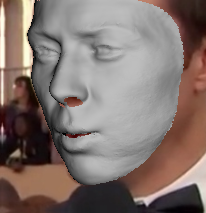}
\includegraphics[width=\suppw\linewidth]{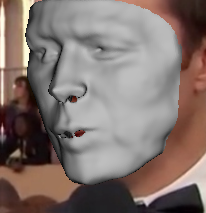}
\includegraphics[width=\suppw\linewidth]{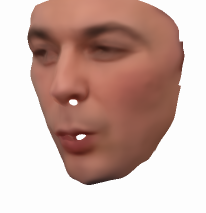}

\includegraphics[width=\suppw\linewidth]{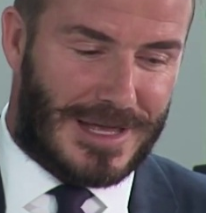}
\includegraphics[width=\suppw\linewidth]{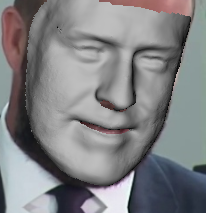}
\includegraphics[width=\suppw\linewidth]{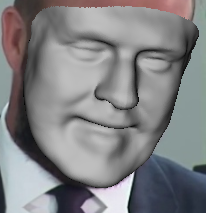}
\includegraphics[width=\suppw\linewidth]{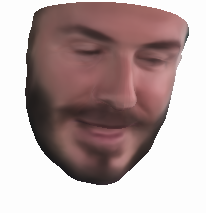}

\includegraphics[width=\suppw\linewidth]{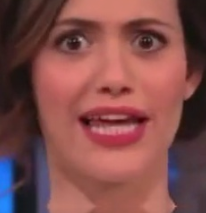}
\includegraphics[width=\suppw\linewidth]{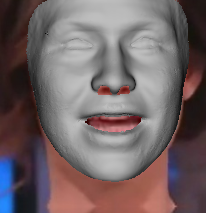}
\includegraphics[width=\suppw\linewidth]{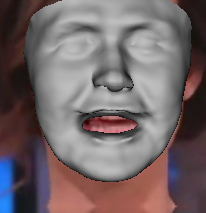}
\includegraphics[width=\suppw\linewidth]{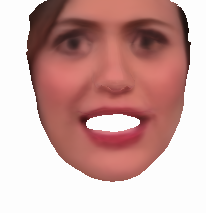}
\hspace{-15pt}

\begin{minipage}{\suppw\linewidth}
    \centering
    \figtext{Input}
\end{minipage}
\begin{minipage}{\suppw\linewidth}
    \centering
    \figtext{DFNRMVS}
\end{minipage}
\begin{minipage}{\suppw\linewidth}
    \centering
    \figtext{Ours}
\end{minipage}
\begin{minipage}{\suppw\linewidth}
    \centering
    \figtext{Ours w/ texture}
\end{minipage}
\end{minipage}}
\subfloat[View 2]{\begin{minipage}{0.495\linewidth}
\includegraphics[width=\suppw\linewidth]{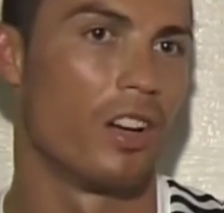}
\includegraphics[width=\suppw\linewidth]{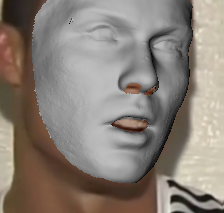}
\includegraphics[width=\suppw\linewidth]{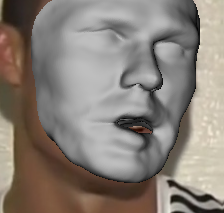}
\includegraphics[width=\suppw\linewidth]{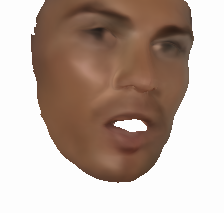}

\includegraphics[width=\suppw\linewidth]{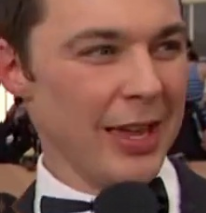}
\includegraphics[width=\suppw\linewidth]{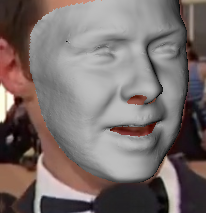}
\includegraphics[width=\suppw\linewidth]{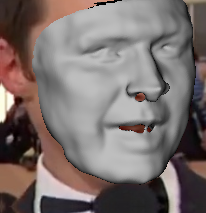}
\includegraphics[width=\suppw\linewidth]{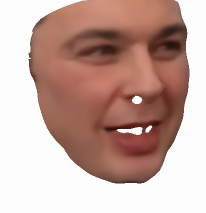}

\includegraphics[width=\suppw\linewidth]{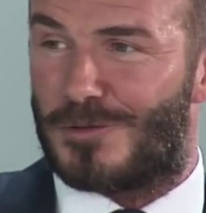}
\includegraphics[width=\suppw\linewidth]{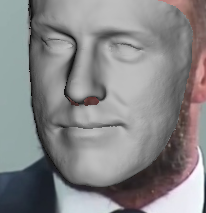}
\includegraphics[width=\suppw\linewidth]{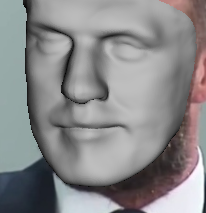}
\includegraphics[width=\suppw\linewidth]{figures/supp/inthewild/s3v0_predimg.png}

\includegraphics[width=\suppw\linewidth]{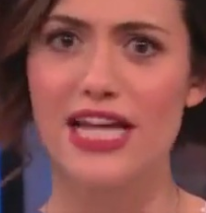}
\includegraphics[width=\suppw\linewidth]{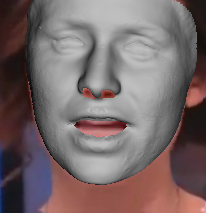}
\includegraphics[width=\suppw\linewidth]{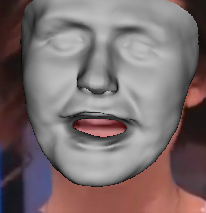}
\includegraphics[width=\suppw\linewidth]{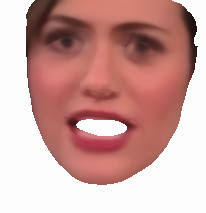}
\hspace{-15pt}

\begin{minipage}{\suppw\linewidth}
    \centering
    \figtext{Input}
\end{minipage}
\begin{minipage}{\suppw\linewidth}
    \centering
    \figtext{DFNRMVS}
\end{minipage}
\begin{minipage}{\suppw\linewidth}
    \centering
    \figtext{Ours}
\end{minipage}
\begin{minipage}{\suppw\linewidth}
    \centering
    \figtext{Ours w/ texture}
\end{minipage}
\end{minipage}}

\vspace{-7pt}
\caption{Qualitative comparison with DFNRMVS \cite{bai20dfnrmvs} based on in-the-wild input images in a two-view setting. Each row presents the input images (\ie, view1, and view2) and the corresponding results of one identity. In each quadruple, the first column is the one of the two input views, the second column is the reconstruction result of DRNRMVS \cite{bai20dfnrmvs}, the third column is our reconstructed mesh, and the last column is our result rendered with reconstructed texture.}
\label{fig:two_view_results}
\end{figure*}
% -----------------------------------

From these two figures, we can see that compared to DFNRMVS~\cite{bai20dfnrmvs}, the geometry of our method is closer to the ground truth with more local details, such as the forehead and the mouth region.
Also, the nose shape of our method is more similar to the input target than that from \cite{bai20dfnrmvs}.
In Figure~\ref{fig:result_gallery_bosp}, the region of the error map is determined by the available ground truth data.

In addition, we qualitatively compare our method with DFNRMVS~\cite{bai20dfnrmvs} in a two-view setting using in-the-wild images of several different identities.
Since \cite{bai20dfnrmvs} outperforms other previous methods \cite{chen2019photo_sv,feng2018joint_sv,tewari2019fml_weak_super_video_mv}, we only present the comparison with \cite{bai20dfnrmvs} in Fig.~\ref{fig:two_view_results}.
\revision{In our optimization-based framework, the entire face is treated equally and the amount of pixels in the eye region is relatively small. Therefore, the color of the reconstructed eyes may be slightly influenced by neighboring pixels of skin. It is possible to alleviate this kind of artifact by introducing different penalty weights for different subregions.
}

\begin{table}%[h!]%[ht]s
\centering
%\fontsize{9pt}{0.9\baselineskip}\selectfont
\begin{tabular}{@{\extracolsep{\fill}} l | c | c c  }
\hline
\hline
Method  & Mean (mm) & STD (mm) \\
\hline
Deng et al. \cite{deng2019accurate_weak-super_mv} & 1.47 & 0.40 \\
DFNRMVS \cite{bai20dfnrmvs} &  1.44 &  0.38\\
Bai et al. \cite{bai2021riggable} & \textbf{1.36} & 0.38 \\
\hline
Ours &  1.39 &  \textbf{0.35}  \\
% Ours & \XSolidBrush &  1.388 &   0.349  \\
\hline
\hline
\end{tabular}
%\vspace{-5pt}
\caption{Quantitative results using two-view input on the Bosphorus dataset \cite{savran2008bosphorus}.}
\label{Table:Comparison_Bosp}
\end{table}

\begin{table}%[h!]%[ht]
\centering
\begin{adjustbox}{width=\columnwidth,center}
%\fontsize{9pt}{0.9\baselineskip}\selectfont
\begin{tabular}{@{\extracolsep{\fill}} l | c | c | c | c }
\hline
\hline
Method & Metric & 2 views & 3 views & 4 views \\
\hline
% \vspace{2pt}
\multirow{2}{*}{DFNRMVS \cite{bai20dfnrmvs}} & {Mean (mm)} & 1.04 &  1.03 & 1.02  \\
% \cline{2-5}
& {STD (mm)} & 0.33 & 0.30 & 0.29 \\
\hline
\multirow{2}{*}{Ours} & {Mean (mm) } & \textbf{0.995} & \textbf{0.991} & \textbf{0.981}  \\
% \cline{2-5}
& {STD (mm)} & 0.177 &  0.206 & 0.196 \\
\hline
\hline
\end{tabular}
\end{adjustbox}
%\vspace{-5pt}
\caption{Quantitative results on the Stirling{\slash}ESRC dataset \cite{stir3d} with different numbers of input views.}
\label{Table:Comparison_Stir_mv}
\end{table}

\subsection{Quantitative Results}
To illustrate the effectiveness of the proposed method, we compare with several state-of-the-art methods quantitatively in multi-view 3D face reconstruction, including \cite{deng2019accurate_weak-super_mv,bai20dfnrmvs,bai2021riggable}.
As shown in Table~\ref{Table:Comparison_Bosp}, our approach achieves better performance in terms of mean errors compared with previous methods \cite{bai20dfnrmvs,deng2019accurate_weak-super_mv} and is comparable to a more recent method \cite{bai2021riggable} when evaluate on the Bosphorus dataset.

We also conduct quantitative comparisons on the test set of Stirling{\slash}ESRC dataset as shown in Table \ref{Table:Comparison_Stir_mv}. Since \cite{deng2019accurate_weak-super_mv, bai2021riggable} have not provided results on this test set, we only compare our results with DFNRMVS \cite{bai20dfnrmvs}.
The results demonstrate that our method has consistently lower mean errors with different numbers of input views.
Moreover, the performance improvement of our method becomes more significant with increased number of views.

Furthermore, it is worth noting that the number of parameters of our method ($4.99$M) is about one order of magnitude smaller than those of existing methods, such as $54.69$M for \cite{bai2021riggable}, $83.21$M for \cite{bai20dfnrmvs}, and $53.35$M for \cite{feng2021learning}.
Overall, we consider our results comparable to \cite{bai2021riggable} while being superior than other SOTA methods in terms of mean errors.

\begin{table}%[t]
\centering
\begin{tabular}{@{\extracolsep{\fill}} l | c | c  }
\hline
\hline
Method & {Mean (mm) } & {STD (mm)}  \\
\hline
%{\makecell[l]{ DFNRMVS \\ \cite{bai20dfnrmvs}}} & 1.04 & 0.33 \\
{DFNRMVS} & 1.040 & 0.33 \\
{DFNRMVS$^{*}$} & 1.171 & 0.35 \\
\hline
Baseline & 1.152 & 0.351 \\
Baseline + view-switch loss & 1.108 & 0.163  \\
Our full algorithm & \textbf{0.995} & 0.177 \\
\hline
\hline
\end{tabular}
%\vspace{-5pt}
\caption{Ablation study on the Stirling{\slash}ESRC dataset \cite{stir3d}. The symbol $^*$ denotes that the results are obtained by testing the released model of DFNRMVS~\cite{bai20dfnrmvs} with the same samples as ours.}
\label{Table:ablation_stir}
\end{table}

\subsection{Ablation Studies}
We perform ablation study on the test set of Stirling{\slash}ESRC dataset to evaluate the impact of the proposed view-switch loss and residual latent code.
The corresponding two-view reconstruction results are shown in Table~\ref{Table:ablation_stir}.
Our baseline is the reconstruction performance obtained by solving the optimization problem as Eq.~\eqref{eqn:test_pca_latentcode_weight}, without using our residual latent code or view-switch loss.

As shown in Table~\ref{Table:ablation_stir}, our proposed view-switch loss leads to better reconstruction performance.
The improvement of our full algorithm with the addition of residual latent code is also noticeable (the last row in Table~\ref{Table:ablation_stir}). This fact demonstrates that our residual latent codes effectively extend the shape space of the morphable models.
Since the Stirling{\slash}ESRC test partition of \cite{bai20dfnrmvs} is not available, we follow the same selection strategy as in \cite{bai20dfnrmvs} and present the test results of their released model on our selected test samples for fair comparison in Table~\ref{Table:ablation_stir}.

\section{Conclusions}
\label{Section:Conclusions}
In this work, we present a novel method for 3D face reconstruction from multi-view images via implicit neural deformation.
By using a neural SDF based representation, we are able to reconstruct faces with high-fidelity details even from sparse-view input of diverse expressions.
Different from previous 3DMM based methods, we propose residual latent code to extend the shape space of implicit morphable models.
To further enforce consistency among different views of one instance at test time, we introduce a novel view-switch loss for joint optimization of the network and latent code.
Besides, we design a training strategy for the implicit neural network to alleviate the collapse issue during self-supervised optimization by introducing prior information of face geometry and colors.
Our results on the Stirling{\slash}ESRC dataset and the Bosphorus dataset demonstrate that our approach outperforms alternative baselines and state-of-the-art methods.

Our current implementation of test-time reconstruction takes about two hours for a single instance and the major bottleneck is the ray-tracing module.
We plan to integrate several recent approaches~\cite{Rebain_2021_CVPR,Garbin_2021_ICCV,tang2021octfield,liu2020neural} for accelerated rendering of neural SDFs to speed up the computation. 

\paragraph{Acknowledgements.}
We would like to thank anonymous reviewers for their valuable feedback and suggestions.

\bibliographystyle{eg-alpha-doi} 
\bibliography{pg22}       

\end{document}